\documentclass[journal]{IEEEtran}
\usepackage{cite}

\usepackage{bbm}
\usepackage{mathtools}
\usepackage{multicol}
\usepackage{array,multirow}
\usepackage{adjustbox}
\usepackage{multicol}
\usepackage{tabularx}
\usepackage{color, colortbl}
\usepackage{hyperref}  
\usepackage{lipsum}
\usepackage{amssymb}
\usepackage{amsthm}
\usepackage{amsmath}
\usepackage{amsfonts}
\usepackage{graphicx}
\usepackage{subfig}
\usepackage{cancel}
\usepackage{algorithm}

\newcommand{\state}{\ensuremath{x}}
\newcommand{\xtraj}{\ensuremath{\mathbf{x}}}

\newcommand{\control}{u}
\newcommand{\utraj}{\ensuremath{\mathbf{u}}}
\newcommand{\uset}{\ensuremath{\mathcal{U}}}

\newcommand{\cost}{C}
\newcommand{\costIdeal}{C^*}
\newcommand{\weight}{\ensuremath{\theta}}
\newcommand{\wspace}{\ensuremath{\Theta}}
\newcommand{\statefeat}{\ensuremath{\phi}}
\newcommand{\trajfeat}{\ensuremath{\Phi}}
\DeclareMathOperator*{\expect}{\mathbb{E}}

\usepackage{soul}
\usepackage{xcolor,cancel}
\setstcolor{red}
\setul{}{0.2ex}

\newcommand{\change}[1]%
    {\textcolor{black}{#1}}    
 
\newcommand{\changenew}[1]%
    {\textcolor{black}{#1}}  

\newcommand{\strike}[1]%
    {}

\usepackage{float}
\usepackage{graphicx}
\ifCLASSINFOpdf
\else
\fi
\usepackage{amsmath}
\usepackage{algorithmic}

\begin{document}
%
\title{Quantifying Hypothesis Space Misspecification \\in Learning from  Human-Robot Demonstrations\\ and Physical Corrections}
%
%
%

\author{Andreea~Bobu,~\IEEEmembership{Student Member,~IEEE,}
        Andrea~Bajcsy,~\IEEEmembership{Student Member,~IEEE,}
        Jaime~F.~Fisac,~\IEEEmembership{Student Member,~IEEE,}
        Sampada~Deglurkar,
        and~Anca~D.~Dragan~\IEEEmembership{Member,~IEEE}
\thanks{Department of Electrical Engineering and Computer Sciences}
\thanks{University of California, Berkeley}
\thanks{Andreea Bobu, UC Berkeley, Berkeley, CA, 94709 USA}
\thanks{e-mail: abobu@eecs.berkeley.edu}
\thanks{A. Bajcsy, J. F. Fisac, S. Deglurkar, and A. D. Dragan are with the EECS Department at University of California, Berkeley.}
\thanks{Manuscript received April 20, 2019; revised \change{October 13, 2019.}}}

%
%

\markboth{IEEE TRANSACTIONS ON ROBOTICS}
{Bobu \MakeLowercase{\textit{et al.}}: Quantifying Hypothesis Space Misspecification in Human-Robot Demonstrations and Physical Corrections}
%


\IEEEspecialpapernotice{(Invited Paper)}


\onecolumn
\thispagestyle{empty}

\begin{center}
This paper has been accepted for publication in \textit{IEEE Transactions on Robotics}.

\bigskip

DOI: \href{https://ieeexplore.ieee.org/document/9007490}{10.1109/TRO.2020.2971415}

IEEE Xplore: \href{https://ieeexplore.ieee.org/document/9007490}{https://ieeexplore.ieee.org/document/9007490}

\bigskip

\end{center}

$\copyright$ 2020 IEEE. Personal use of this material is permitted.  Permission from IEEE must be obtained for all other uses, in any current or future media, including reprinting/republishing this material for advertising or promotional purposes, creating new collective works, for resale or redistribution to servers or lists, or reuse of any copyrighted component of this work in other works.

\clearpage\setcounter{page}{1}

\twocolumn

\maketitle



\begin{abstract}

Human input has enabled autonomous systems to improve their capabilities and achieve complex behaviors that are otherwise challenging to generate automatically. 
Recent work focuses on how robots can use such input --- like demonstrations or corrections --- to learn intended objectives. These techniques assume that the human's desired objective already exists within the robot's hypothesis space. 
In reality, this assumption is often inaccurate: there will always be situations where the person might care about aspects of the task that the robot does not know about. Without this knowledge, the robot cannot infer the correct objective. 
Hence, when the robot's hypothesis space is \textit{misspecified}, even methods that keep track of uncertainty over the objective fail because they reason about which hypothesis might be correct, and not whether \emph{any} of the hypotheses are correct.
In this paper, we posit that the robot should reason explicitly about how well it can explain human inputs given its hypothesis space and use that \textit{situational confidence} to inform how it should incorporate human input. We demonstrate our method on a 7 degree-of-freedom robot manipulator in learning from two important types of human input: demonstrations of \strike{manipulation} \change{motion planning} tasks, and physical corrections during the robot's task execution.

\end{abstract}


%
\IEEEpeerreviewmaketitle

\section{Introduction}
\label{sec:intro}

%
%
%
%

\begin{figure}[t]
    \centering
    \includegraphics[width=\columnwidth]{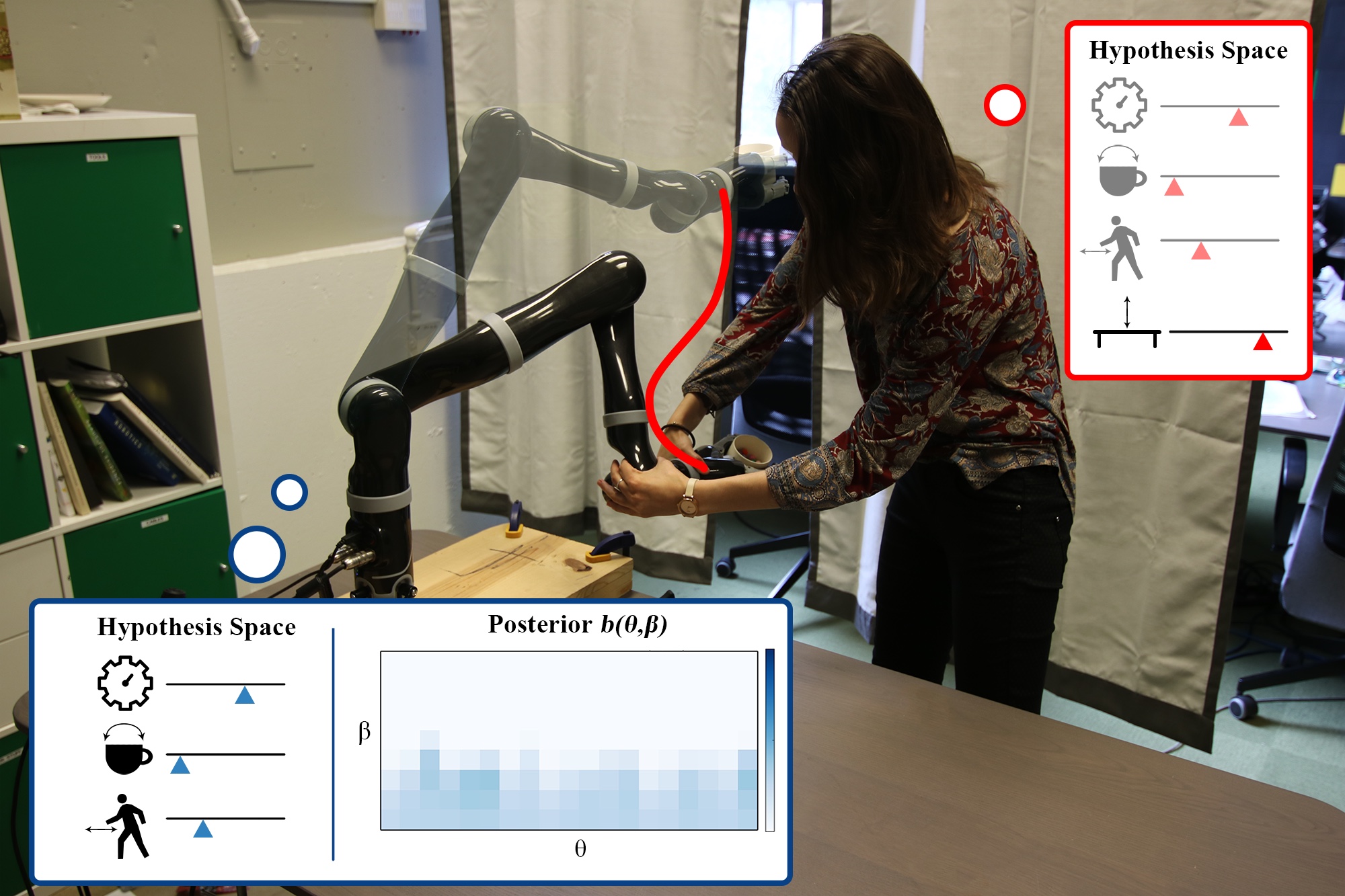}
    \caption{A household robotics scenario where the person physically interacts with the robot. The person prefers the robot to keep cups closer to the table, but accounting for the table (outside of collisions) is not in the robot's hypothesis space for what the person might care about. Thus, the robot's internal situational confidence, $\beta$, about what the human input means is low for all hypotheses $\theta$.}
    \label{fig:front_fig}
\end{figure}

\IEEEPARstart{A}{utonomous} systems are increasingly interfacing and collaborating with humans in a variety of contexts, such as semi-autonomous driving, automated control schemes on airplanes, or household robots working in close proximity with people. While the improving capabilities of robotic systems are opening the door to new application domains, the substantially greater complexity and interactivity of these settings makes it challenging for system designers to account for all relevant operating conditions and requirements ahead of time. For example, a household robot designer may not know how an end-user would like the robot to interact with the personal possessions in the user's home.

In situations like these, it can be beneficial for the robot to utilize human input as guidance on the desired behavior. In fact, human input has enabled researchers and engineers to program advanced behaviors that would have otherwise been extremely challenging to specify. Helicopter acrobatics \cite{abbeel2010autonomous}, aggressive automated car maneuvers \cite{kolter2010probabilistic}, and indoor navigation \cite{kuderer2012feature} are three cases that exemplify the benefit of using human input for guiding robot behavior. 

In order to utilize human input, system designers typically equip robots with a representation of possible objectives that the human \emph{could} care about. These representations can range from quadratic cost models\cite{kopf2017irllqr} to complex temporal logic specifications \cite{fu2015pareto} to neural networks \cite{finn2016guided}. However, anticipating all motivations for human input and specifying a complete model is challenging\strike{ for system designers}. Consider Fig. \ref{fig:front_fig} where a human is attempting to change the robot's behavior in order to make it consistently stay close to the table, but the robot's model of what the human might care about does not include distances to the table. 
By choosing a class of functions, the system designer implicitly assumes that what the human wants (and is giving input about) can be represented via a member of that class. Unfortunately, when this assumption breaks, the system can misinterpret human guidance, perform unexpected or undesired behavior, and degrade in overall performance.

Two approaches to mitigating this problem could be to either start with a more complex objective space or to continuously increase its complexity given more data. Unfortunately, even complex models are not guaranteed to encompass all possibilities and re-computing the best objective space based on human data faces the threat of overfitting to the most recent observations.
In contrast, we argue the robot should be able to \textit{understand when it cannot understand the input}. For example, if the end-user in the home is trying to guide the robot to handle fragile objects with care but the system does not posses a model of fragility, the robot should deduce that this input cannot be well explained by any of its given hypotheses. 

In this work, we formalize how autonomous systems can explicitly reason about how well they can explain given human inputs. To do this, we observe that \strike{a human input will appear unlikely with respect to all possible hypotheses if the robot's model is misspecified} 
\change{if a human input appears unlikely with respect to all possible hypotheses, then the robot's model is misspecified}. We build on previous work centered around this observation to propose a Bayesian inference framework focused on inferring both model parameters, and their corresponding \textit{situational confidence}. If the robot is in situations like Fig. \ref{fig:front_fig} where none of the hypotheses explain the human's input well, then the situational confidence will be low for all hypotheses, indicating that the robot's model is not sufficiently rich to understand the human's input. However, when the robot's model is well specified, our framework does not impede the robot from inferring the correct task objectives --- in fact, the situational confidence will be high, providing an indicator of how well the system can understand the objective.

We illustrate the utility of situational confidence estimation \change{in quantifying objective space misspecification} for two types of human input: demonstrations and corrections. \strike{Although we highlight the utility in these two domains, the principles outlined in our formalism are more general and have implications for a wider range of interaction modes, including shared autonomy and autonomous driving.} 
Our contributions in this work are: 
\begin{enumerate}
    \item \strike{formulate the problem of situational confidence estimation}
    \change{we introduce a general framework for quantifying objective space misspecification} when the human and the robot are acting on the same dynamical system;
    \item \strike{instantiate this problem formulation for two domains -- learning from demonstrations and learning from physical human-robot corrections, and propose algorithms for tractable inference}
    \change{we showcase the framework for learning from demonstrations using user demonstration data for an arm motion planning task;}
    \item \strike{conduct an in-person user study of our approach on a 7-DoF robot manipulator with 12 human participants.}
    \change{we showcase the framework for learning from physical corrections by deriving an algorithm for online (close to real-time) inference and testing it in a user study.}
\end{enumerate}

We note that this work is an extension of \cite{bobu2018learning}, which was originally presented at the Conference on Robot Learning, 2018. \change{We build on this work by introducing a general framework for quantifying objective space misspecification, and instantiating it in a new type of human input: learning from demonstrations. Not only are demonstrations the most widely used type of input for learning objective functions, but the applicability across two input types suggests that the approach could be adapted more broadly to more types of human feedback.}

The remainder of this paper is organized as follows:
Section \ref{sec:related} places this work in the context of existing literature on robots learning from humans and model confidence estimation. 
Section \ref{sec:problem} frames the confidence estimation problem more formally for scenarios where the human and robot operate on the same dynamical system.
\strike{Sections \mbox{\ref{sec:demonstrations}} and \mbox{\ref{sec:corrections}} present algorithmic approaches to estimating situational confidence in the learning from demonstration and from human corrections domains.} \change{Section \ref{sec:demonstrations} directly instantiates the framework in Section \ref{sec:problem} for the case of learning from demonstrations.}
\change{Section \ref{sec:corrections} presents a derivation of approximations of the general formalism for tractable online inference from human corrections.}
Section \ref{sec:case_study} showcases our proposed approach in several case studies where the robot's hypothesis space cannot or only partially explain the human's input.
Section \ref{sec:study} presents the results of a user study of our approach as applied to a 7-DoF robotic manipulator learning from human participants. 
Section \ref{sec:conclusion} concludes with a discussion of some of the limitations of our work, as well as suggestions for future research directions.

Overall, we think that the ability to detect misspecification when learning objectives from human input will become increasingly important as robotics capability advances and we will want end-users to customize how the robot behaves. Our work takes a step in this direction by enabling robots to detect when none of the hypotheses they have explain the user input, and our experiments show promising results. Of course, there are still limitations to this. \change{One limitation is in the experiments themselves, which are only for motion planning tasks with low-dimensional hypothesis spaces. A more fundamental limitation is that} there will still be cases when the person wants something outside the robot's hypothesis space, but the robot can nonetheless explain their current input relatively well with what it has access to, thus confusing misspecification for slight noise in the human input. \change{This will especially be the case as the hypothesis space is more expressive, and can only be solved by the robot receiving a lot more human input: each might be explainable by some hypothesis, but eventually no hypothesis can explain all input.} More work is needed in studying how to \change{query for diverse human input, as well as how to }convey what the robot has learned back to the person, and in general how to have a true collaborative interaction to detect and resolve misspecification in the objective space.

\section{Related Work}
\label{sec:related}
We group prior work into three main categories: enabling robots to learn from human input, doing so while leveraging uncertainty, and estimating model confidence.

\subsection{Robots learning from humans}

The programming of robots through direct human interaction is a well-established paradigm. Human input can be given to the robot in a variety of forms, from teleoperation of the robot by a user to kinesthetic teaching \cite{argll2009survey}.

In such interaction paradigms, the robot aims to infer a \textit{cost function} or \textit{policy} that best describes the examples that it has received. New avenues of research focus on learning such robot objectives from human input through demonstrations~\cite{abbeel2004apprenticeship,osa2018algorithmic}, teleoperation data~\cite{javdani2015shared}, corrections~\cite{jain2015learning,bajcsy2017phri}, comparisons~\cite{christiano2017preferences}, examples of what constitutes a goal~\cite{fu2018variational}, or even specified proxy objectives~\cite{HadfieldMenell2017InverseRD}. In this paper, we focus on learning from two of such types of human input -- demonstrations and physical corrections -- although we stress that the principles outlined in our formalism are more general and could be applied to the other interaction modes mentioned.

One approach to learning behaviors from human inputs is inverse reinforcement learning (IRL). In classical IRL, the robot receives complete optimal \textit{demonstrations} of how to perform a task, and the robot learns the human's cost function from these observations \cite{Kalman1964inverse, Ng2000inverse, osa2018algorithmic}. In this paradigm, it is typically assumed that the expert is trying to optimize an unknown cost function. The robot uses the observations of the human's behavior to recover the underlying objective.

Another useful form of human input are \textit{corrections}: here, the robot performs the task according to how it was programmed and the user corrects aspects of the task to better match their preferences. From these sparse interactions, the robot also performs cost function inference to improve performance during the next task iteration \cite{shivaswamy2015coactive, ratliff2006maximum, karlsson2017autonomous}. Examples of learning from corrections have been explored in offline~\cite{jain2015learning},\change{\cite{Gutierrez2018IncrementalTM}} and online settings~\cite{bajcsy2017phri, HRI},\change{\cite{Celemin2019,Argall2011tactile}}. 

Although powerful, the aforementioned IRL works assume that the human expert provides optimal demonstrations, which is often an unrealistic assumption. Real human input, especially during interaction with high degree-of-freedom systems like robotic manipulators, is noisy and sub-optimal. Second, much of the corrections literature has focused on estimates of the human's objectives. However, in practice, even the most likely estimate might not be a very likely one. Thus, in both domains, we stress that it is important to maintain the uncertainty over the estimated objectives.

\subsection{Uncertainty in robot learning}

Rather than estimating a single objective, some learning methods maintain an entire probability distribution over what the objective might be \change{\cite{brown2018risk,HadfieldMenell2017InverseRD, losey2018including,ramachandran2007bayesian}}. This not only enables the robot to leverage a prior, but also to then generate its behavior in a way that is mindful of the entire distribution, rather than just using the the maximum likelihood estimator\strike{, increasing its robustness}.

Bayesian IRL \cite{ramachandran2007bayesian} treats demonstrations as evidence about the objective, and does a Bayesian belief update on a prior distribution. Inverse Reward Desing \cite{HadfieldMenell2017InverseRD} treats the objective a designer specified for a particular set of environments (a ``proxy'' objective) as evidence about the true desired objective, again obtaining a full distribution over what the designer might actually want.
The intuition is that this observed proxy objective (that may be misspecified) incentivizes behavior that is approximately optimal with respect to the true  objective.

Lastly, specifically for input as physical corrections, \cite{losey2018including} reasons over the uncertainty of the estimated human preferences through the means of a Kalman filter. The method maintains a mean estimate and a covariance of this estimate as a measure of confidence. These are used in planning the robot's trajectory such that it optimizes for features it is confident about, while avoiding features it is uncertain about.

Although they maintain a full distribution, these works still assume that what the human wants is in the robot's objective space. 
We argue that this is not necessarily a realistic assumption, and later showcase some consequences that arise when it is not true. When the robot's hypothesis space is misspecified, even when maintaining uncertainty over the objective, state-of-the-art methods interpret human input as evidence about which hypothesis is correct, rather than considering whether any hypothesis is correct. In this work, we focus on the latter.

\subsection{Situational confidence estimation}

Some recent works are studying how to enable robots to understand that their models cannot explain human input well\change{~\cite{Zheng2014robustBIRL,fisac2018probabilistically,fridovich-keil2019confidence}}. The authors in \cite{fisac2018probabilistically,fridovich-keil2019confidence} employ a noisily-optimal model of human pedestrian motion when the human and the robot operate on separate dynamical systems (and have separate objective functions). The paper introduces the notion of model confidence estimation and uses the apparent likelihood of the human's choice of actions to adjust the confidence in predictions about their behavior. 

This work draws inspiration from the notion of model confidence estimation, generalizing it to the setting of inferring what the robot's objective ought to be. Instead of focusing on misspecification of a discrete set of physical goal locations for pedestrian navigation, here we study misspecification of a relatively complex set of possible robot objectives in \strike{manipulation} \change{motion planning} tasks. As a result of focusing on robot objectives, we also study a different form of human input -- that is, input in the context of operating on the same dynamical system, such as full task demonstrations and physical corrections.

\section{Problem Formulation and Approach}
\label{sec:problem}

We consider a robot $R$ operating in the presence of a human $H$ whom it seeks to assist in the execution of some task. In the most general setting, the robot and the human are both able to affect the evolution of the state $\state \in \mathbb{R}^n$ over time through their respective control inputs:
    \begin{equation}\label{eq:dynamics}
        \state^{t+1} = f\left(\state^t,\control_R^t,\control_H^t\right)
        \;,
    \end{equation}
with $\control_R \in \uset_R$ and $\control_H\in \uset_H$, where $\uset_i$ ($i\in\{H,R\})$ are compact sets.
We assume that the human has some consistent preference ordering between different state trajectories and input signals, which could in principle be expressed through a cost function of the form
\begin{equation}\label{eq:abstract_cost}
    \costIdeal(\xtraj, \utraj_R, \utraj_H)
\end{equation}
where the state trajectory is $\xtraj = [x^0, x^1, \hdots, x^T]\in\mathbb{R}^{n(T+1)}$, the robot's control input is $\utraj_R = [u_R^0, u_R^1, \hdots, u_R^T]\in\mathbb{R}^{n(T+1)}$, and the human's is $\utraj_H = [u_H^0, u_H^1, \hdots, u_H^T]\in\mathbb{R}^{n(T+1)}$.%
\footnote{%
For deterministic dynamics \eqref{eq:dynamics}, having~$\state^0, \utraj_R$ and~$\utraj_H$ is enough to fully specify the entire state trajectory~$\xtraj$. In this case, the cost function could be rewritten as $\costIdeal(\state^0, \utraj_R, \utraj_H)$ by implicitly encoding \eqref{eq:dynamics}. For clarity, we use the more general form in \eqref{eq:abstract_cost} and make the dependence explicit where needed.
}
Note that this hypothesized cost function~$\costIdeal$ can be quite general, encoding an arbitrary preference ordering. However, the robot does not in general have access to the human's preferences $C^*$, and must instead attempt to infer and represent them tractably. 

In order to do this, the robot can typically reason over a parametrized approximation of the cost function, which introduces an inductive bias, making inference tractable at the cost of limiting expressiveness: 
in some cases, the chosen set of parametric functions may fail to encode preferences that would explain the human's behavior with sufficient accuracy. 
In this work, we will denote by~$\cost_{\weight}$ the cost function induced by parameters~$\weight \in \wspace $, and the robot seeks to estimate the human's preferred~$\weight$ from her control inputs~$\utraj_H$.

In a general setting, since the state trajectory~$\xtraj$ is determined not only by the human's actions~$\utraj_H$ but also the robot's~$\utraj_R$, the human would need to reason about how the robot will respond to her decisions.
This requires analyzing the interaction in a game-theoretic framework \cite{Hadfield-Menell2016CIRL, Fisac2017PragmaticPedagogicVA}, which will not be the object of this work.
Instead, we focus on common interaction scenarios in which the robot can approximately assume that the human does not explicitly account for the coupled mutual influence between both agents' decisions.
This happens frequently if the human is either providing a demonstration for the robot or intervening to correct the robot's default behavior.
In these settings, the typical assumption is that the human has all necessary information about the robot's control input~$\utraj_R$ before deciding on her own~$\utraj_H$.

Thus, given observations of the human input $\utraj_H$ from an initial state $\state^0$,
the robot needs to draw inferences on the cost parameter~$\weight$: 
    \begin{equation}\label{eq:theta_posterior}
        P(\weight \mid \state^0, \utraj_R, \utraj_H) = \frac{P(\utraj_H \mid \state^0, \utraj_R; \weight) P(\weight)}{ \int_{\bar{\weight}} P(\utraj_H \mid \state^0, \utraj_R; \bar{\weight}) P(\bar{\weight}) d\bar{\weight}}
        \enspace,
    \end{equation}
where $P(\utraj_H \mid \state^0, \utraj_R; \weight)$ characterizes how the robot expects the human's input to be informed by her preferences, conditioned on the initial state and the robot's expected controls.

For example, if the human were assumed to \strike{always }act optimally, this model would place all probability on the set of \strike{strictly  }optimal states and actions with respect to the cost $\cost_{\weight}$. Of course, this would be an unreasonably strong assumption given that the robot's parametrized cost constitutes a best effort to approximate the human's preferences. Instead, a useful modeling choice can be to  characterize the human as being more \emph{likely} to take actions that are well-aligned with her preferences. 

\change{One such model is inspired by the Boltzmann energy-based model satisfying the maximum entropy principle~\cite{jaynes1957infotheory}.} Following \change{its adaptations as a model of human decision-making in} ~\cite{von1945theory,baker2007goal,bajcsy2017phri}, we model the human as a noisily-optimal agent that tends to choose control inputs that approximately minimize the modeled cost:
\begin{equation}
\begin{aligned}
    P(\utraj_H \mid \state^0, \utraj_R; \weight,\beta)
    =& \frac{
        e^{-\beta \cost_{\weight}\big(\xtraj(\cdot;\state^0,\utraj_R,\utraj_H),\utraj_R, \utraj_H\big)}
    }{
        \int_{\bar{\utraj}_H}e^{-\beta \cost_{\weight}\big(\xtraj(\cdot;\state^0,\utraj_R,\bar\utraj_H),\utraj_R, \bar\utraj_H\big)} d\bar{\utraj}_H
    }. 
  \label{eq:boltzmann}
\end{aligned}
\end{equation}
In this model, the inverse temperature coefficient $\beta\in [0,\infty)$ determines the degree to which the robot expects to observe human actions that are consistent with the cost model. 

The goal is to detect when the robot does not have a rich enough hypothesis space, i.e. when $\costIdeal$ lies far outside of any $\cost_{\weight}$. 
We call this problem \textit{objective space misspecification}. Rather than only interpreting human input as evidence about \emph{which} hypothesis is correct, we additionally focus on considering whether \emph{any} hypothesis is correct. It is thus crucial that the robot can quantify the extent to which any parameter value $\weight\in\wspace$ can correctly explain the observed human input. 



\subsection{Situational confidence estimation}

The key to our approach goes back to the inverse temperature parameter  $\beta$ in \eqref{eq:boltzmann}. Typically, $\beta$ is a fixed term, encoding the degree to which the robot expects to observe human actions that are optimal. Setting it to 0 models a randomly-acting human, while setting it to $\infty$ models a perfectly optimal human. However, the possibility of objective space misspecification brings fixing $\beta$ into question: when the space is correctly specified, we would expect the human actions to indeed be somewhat close to optimal; but when the space is misspecified, \emph{we should expect the actions to be far from optimal for any $\weight$.} Thus, rather than treating $\beta$ as a fixed term, we build on the work in
\cite{fisac2018probabilistically,fridovich-keil2019confidence}
and explicitly reason over $\beta$ as an additional inference parameter along with $\weight$.
Since $\beta$ directly impacts the entropy of the human's decision model, it can be used as an effective and computationally efficient measure of the robot's confidence in its parametric interpretation of the human's preference:
we say that the robot is assessing its \textit{situational confidence} for the inference task at hand. 

Thus, the robot maintains a joint Bayesian belief  $b(\weight, \beta)$.
For each new measurement of $\utraj_H$ given $\state^0, \utraj_R$, this belief is updated as:
    \begin{equation}
    \label{eq:bayes_update}
        b'(\weight, \beta) = \frac{P(\utraj_H \mid \state^0, \utraj_R; \weight, \beta)b(\weight, \beta)}{\int_{\bar{\weight}, \bar{\beta}}P(\utraj_H \mid \state^0, \utraj_R; \bar{\weight}, \bar{\beta})b(\bar{\weight}, \bar{\beta})d\bar{\weight}d\bar{\beta}}
        \enspace,
    \end{equation}
where $b'(\weight, \beta) = P(\weight, \beta \mid \state^0, \utraj_R, \utraj_H)$. 

This inference can be seen as analogous to performing \change{Bayesian Inverse Reinforcement Learning~\cite{ramachandran2007bayesian} with the} Maximum Entropy Inverse Optimal Control \cite{maxent} \change{observation model}, where we \change{maintain the full belief instead of just the maximum likelihood estimate, and we} explicitly reason over the additional scaling parameter $\beta$. By actively performing inference over $\beta$, the robot can gain insight into the reliability of its human model in light of new evidence. 

\subsubsection{\changenew{Context-dependent usage of situational confidence}}

\changenew{How this insight should be used is dependent on the context of the robot's operation. Here, we provide some examples of how situational confidence can be integrated into various human-robot interaction scenarios and robot motion planners.}

\changenew{In collaborative settings where the human and robot are accomplishing a task together (e.g. manipulating an object together), it may be desirable for the robot to stop and ask for clarification from the human whenever sufficient probability mass indicates low confidence:}
\changenew{
\begin{equation}
    \label{eq:threshold_behavior}
    \forall \weight\in\wspace, \arg\max_\beta b'(\beta \mid \weight) < \epsilon
    \enspace.
\end{equation}
}
\changenew{That is, for a predefined threshold $\epsilon$, if all hypotheses have the most mass on $\beta$s lower than $\epsilon$, the robot can raise a flag.}

\changenew{In assistive applications, where the robot is carrying out a task in close physical proximity to the human, the robot may receive intermittent human input to correct it's task performance. In such scenarios, it may be appropriate for the robot to simply dismiss human corrections that it cannot explain in terms of modeled preference parameters and carry on with its pre-defined task. That is, when a human input results in a $b'(\weight,\beta)$ that satisfies \eqref{eq:threshold_behavior}, the input gets discarded.}

\changenew{Situational confidence could also be leveraged by robot motion planners that excel at decision making under uncertainty. Here, the robot may use its joint posterior belief $b'(\weight,\beta)$ to make goal-driven decisions in the presence of the human. To this end, the coupling between the inference problem and the robot's planning problem can be viewed as a partially observable Markov decision process (POMDP), where the hidden parts of the state are the cost parameter $\theta$ and the situational confidence $\beta$, the robot receives observations about them via human actions $\utraj_H$, it takes actions $\utraj_R$, and it optimizes an unknown parametrized cost $\cost_\weight$. Our problem is, thus, akin to identifying misspecification in the state space of the POMDP. However, inference and planning in such spaces requires solving the full POMDP, which is computationally intractable for large, real-world problems \cite{kaelbling1998pomdps}.}

\changenew{Alternative, less computationally demanding motion planning approaches are also amenable to our framework, where the robot plans to minimize the expected cost for the human given its current belief, by marginalizing over $\beta$:
\begin{equation}
    \label{eq:optimize_without_confidence}
    \min_{\utraj_R} \expect_{\weight\sim b} \big[ \cost_\weight(\xtraj,\utraj_R,\utraj_H)\big]
    \enspace,
\end{equation}
for an expected human input $\utraj_H$ that will typically be~$\mathbf{0}$ if the robot is attempting to successfully perform the task without the need for active human intervention.
To understand the implication \eqref{eq:optimize_without_confidence} has as a function of the inference over $\beta$, we need to understand the posterior belief marginalized over $\beta$ that we are taking the expectation over. At one extreme, if for all $\weight$s the conditional distribution $b'(\beta \mid \theta)$ puts all probability mass on $\beta=0$ (i.e. input poorly explained), since $P(\utraj_H \mid \state^0, \utraj_R; \weight,\beta=0)$ is the same for all $\weight$s, the robot will obtain a posterior for $\weight$ that is equal to the prior. The optimization above becomes the same as optimizing using the robot's prior, i.e. the robot ignores the human input. At the other extreme, if there is one $\weight$ that perfectly explains the input and all others do not, the posterior will put all probability mass on that $\weight$, and the robot will switch to optimizing it.}

\changenew{The objective expectation may also be appropriately weighted by the robot's situational confidence for each $\weight$:
\begin{equation}
    \label{eq:optimize_with_confidence}
    \min_{\utraj_R} \expect_{\weight,\beta\sim b} \big[\beta \cost_\weight(\xtraj,\utraj_R,\utraj_H)\big]
    \enspace,
\end{equation}
which leads to the robot prioritizing those components of the task about which it is most certain.}

\changenew{In Sections \ref{sec:demonstrations} and \ref{sec:corrections} we discuss some of these possibilities in the context of learning from demonstrations and corrections.}

\subsection{Cost representation through basis functions} \label{sec:formulation_cost}


One way to approximate the infinite-dimensional space of possible cost functions using a finite number of parameters is the use of a finite family of basis functions $\trajfeat_i$\cite{Ng2000inverse}. This family can be seen as a truncation of an infinite collection of basis functions spanning the full function space.
Parametric approximations $\cost_\weight$ of the cost function $\costIdeal$ then have the form
\begin{equation}
    \label{eq:traj_features}
    \cost_\weight(\xtraj, \utraj_R, \utraj_H) =
    \sum_{i=1}^d \weight^i \trajfeat_i(\xtraj, \utraj_R, \utraj_H) = \weight^T\trajfeat(\xtraj, \utraj_R, \utraj_H)
    \enspace.
\end{equation}
Consistent with classical utility theories \cite{von1945theory}, we further assume that the human's preferences can be approximated through a cumulative return over time, rewriting \eqref{eq:traj_features} as
\begin{equation}
\label{eq:linear_cost}
        \cost_{\weight}(\xtraj, \utraj_R, \utraj_H) = 
        ~\sum_{i=1}^d \weight^i \sum_{t=0}^T \statefeat_i(\state^t, \control_R^t, \control_H^t) 
        \enspace,
\end{equation}
where $\statefeat_i:\mathbb{R}^n \times \uset \times \uset \rightarrow \mathbb{R}$ are fixed, pre-specified, bounded real-valued basis functions, $\weight$ is the unknown parameter that the robot is trying to fit according to the human's preferences, and $d$ is the dimensionality of its domain $\wspace$.

In the domains presented in Sections \ref{sec:demonstrations} and \ref{sec:corrections}, the functions~$\statefeat_i$ output feature values that encode key aspects of a task---for example distance between the robot body and obstacles in the environment, speed of the motion, or characteristics of a \strike{manipulation} \change{motion planning} task.
In general, the~$\statefeat_i$ can either be hand-engineered by a system designer or more generally learned through data-driven approaches \cite{finn2016guided}.
   
It is important to stress that the misspecification issue we are trying to mitigate is quite general and does not exclusively affect objectives based on hand-crafted features: any model could ultimately fail to capture the underlying motivation of some human actions. While it may certainly be possible, and desirable, to continually increase the complexity of the robot's model to capture a richer space of objectives, there will still be a need to account for the presence of yet-unlearned components of the true objective. In this sense, our work is complementary to open-world objective modeling efforts.  
    
Note that, using a cost model in the form of \eqref{eq:linear_cost}, the observation model \eqref{eq:boltzmann} becomes overparametrized, since for any $(\weight,\beta)$ pair with $\weight \in \wspace$ and $\beta \in [0,\infty)$, one can always find a different $\weight' = c\weight$ with an associated $\beta' = \beta/c$ leading to the same probability distribution over human choices. \change{This is equivalent to using an unrestricted $\wspace$ and $\beta=\|\weight\|$.}
Due to this overparametrization, the absolute value of $\beta$ does not have a universal meaning, and restricting~$\weight$ to have a fixed norm is necessary in order to make comparisons between the $\beta$ values associated to different $\weight$ hypotheses. We thus restrict our~$\wspace$ to the set of vectors with unit norm.

Consider the case where the human provides input \strike{about a feature}\change{for a cost function} in the robot's objective space. This results in the robot inferring high probability on the corresponding $\weight$ vector on the unit sphere with a high magnitude $\beta$. However, if the \strike{feature}\change{cost} that the human cares about and provides input for is outside the robot's hypothesis space, the robot will infer low probability on all $\weight$ vectors in the unit sphere, with low magnitude $\beta$s. 

We now proceed by describing the explicit algorithmic approaches to inferring situational confidence in the learning from demonstrations and corrections domains.

\section{Algorithmic Approach: Demonstrations}
\label{sec:demonstrations}

\begin{figure}[t!]
\centering
\includegraphics[width=\columnwidth]{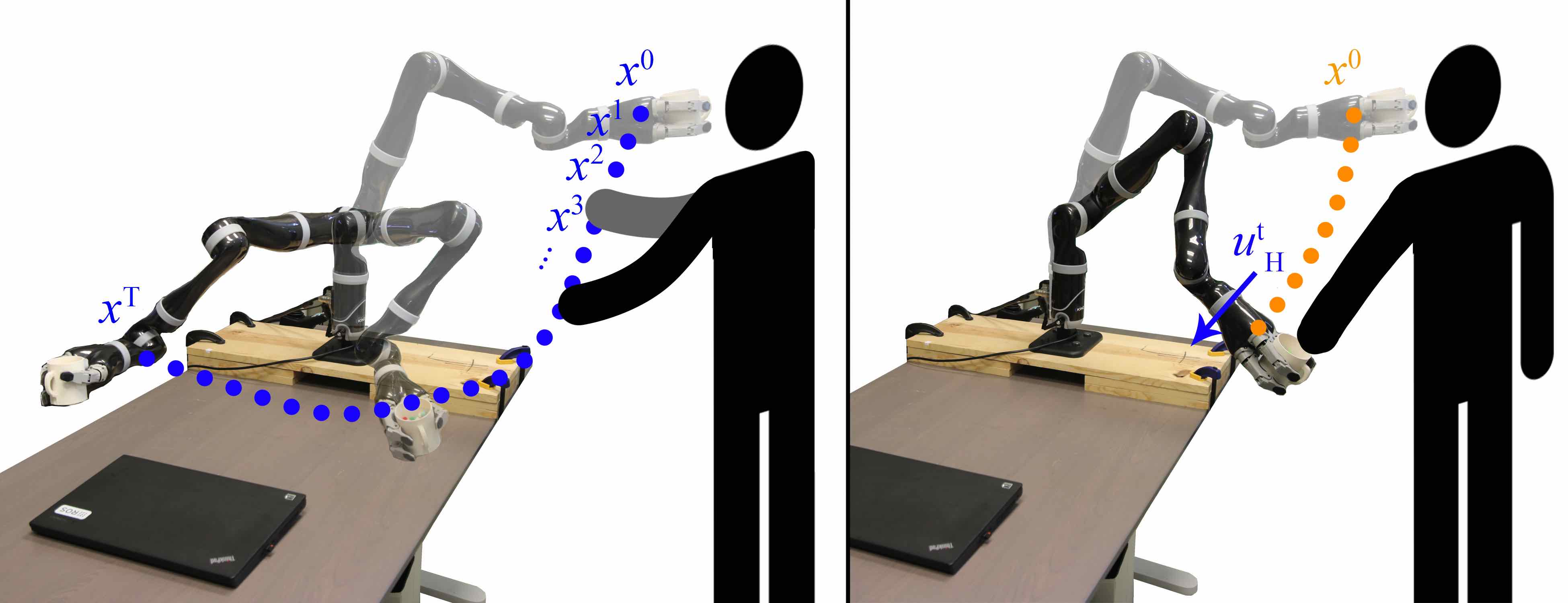}
\caption{(Left) Visual example of a full human-provided demonstration $\xtraj$. (Right) Visual example of a human physical correction $\control^t_H$ onto the robot's current trajectory $\xtraj$.}
\label{fig:demos_and_corrections}
\end{figure}

\subsection{Formulation}

In learning from demonstrations, the human directly controls the state trajectory $\xtraj$ through her input $\utraj_H$, which enables her to offer the robot a demonstration of how to perform the task. Fig. \ref{fig:demos_and_corrections} (left) is an example of such a demonstration.

During the demonstration, the robot is often put in gravity compensation mode or is teleoperated, to grant the person full control over the desired trajectory. As such, in this setting, the cost function $\cost_\weight$ does not depend on the robot controls $\utraj_R$. Additionally, since the person is primarily concerned with \change{the robot's states and not with the (robot or human) actions required to reach those states} \strike{the quality of the final demonstrated trajectory}, we model the human's internal preferences as only dependant on the \strike{final} state trajectory \change{$\xtraj$}. 
Accordingly, the cost function in \eqref{eq:linear_cost} becomes:
\begin{equation}
    \begin{aligned}
        \cost_{\weight}(\xtraj) = \weight^T\trajfeat(\xtraj).
    \label{eq:demo_cost}
    \end{aligned}
\end{equation}
\change{The cost does not have a direct dependence on the actions, but it has an indirect one, as $\xtraj$ depends on $\utraj_R$ and $\utraj_H$.}

In our problem formulation, we would like the robot to explicitly reason about how well it can explain the demonstration given its human model. Thus, we can adapt the model in \eqref{eq:boltzmann} to use this new cost function\change{\footnote{\change{For deterministic \eqref{eq:dynamics}, $P(\utraj_H \mid \state^0, \utraj_R; \weight,\beta)$ is equivalent to $P(\xtraj \mid \weight,\beta)$.}}}, 
\change{
\begin{equation}
    \begin{aligned}
     P(\xtraj \mid \weight,\beta) =& \frac{e^{-\beta \weight^T \trajfeat(\xtraj)}}{\int_{\bar{\xtraj}}e^{-\beta \weight^T \trajfeat(\bar{\xtraj})}d\bar{\xtraj}}
     \enspace ,  
    \label{eq:demo_boltzmann}
   \end{aligned}
\end{equation}
}
then perform the Bayesian update in \eqref{eq:bayes_update}
\change{
\begin{equation}
         b'(\weight, \beta) = \frac{P(\xtraj \mid \weight,\beta)b(\weight,\beta)}{\int_{\bar{\weight},\bar{\beta}}P(\xtraj \mid \bar{\theta},\bar{\beta})b(\bar{\theta},\bar{\beta})d\bar{\theta}d\bar{\beta}}
         \enspace.
 \label{eq:posterior_demo}
\end{equation}
}
\change{Given $b'(\weight,\beta)$, we now can use any of \eqref{eq:threshold_behavior}, \eqref{eq:optimize_without_confidence} or \eqref{eq:optimize_with_confidence}.}
Next, we discuss \strike{practical considerations for }making inference with \eqref{eq:demo_boltzmann} and \eqref{eq:posterior_demo} tractable.

\subsection{Approximation}

Although the proposed formalism enables us to capture if the robot's hypothesis space cannot explain the human's input, it is non-trivial to implement tractably for continuous $\beta$ and $\weight$, and large state and action spaces. Concretely, notice that equations \strike{\mbox{\eqref{eq:boltzmann}} and \mbox{\eqref{eq:bayes_update}}} \change{\eqref{eq:demo_boltzmann} and \eqref{eq:posterior_demo}} constitute a doubly-intractable system with denominators that cannot be computed exactly. \change{For this reason, we employ several approximations in order to demonstrate the benefits of estimating situational confidence. Note that we do not consider these a contribution of our work: we choose the simplest approximations that facilitate tractability. There are many methods for approximate inference of $\theta$ studied in the literature that could be used for the joint ($\theta,\beta$) spaces as well, from Metropolis Hastings~\cite{HadfieldMenell2017InverseRD,Sadigh2017ActivePL}, to acquiring an MLE only via importance sampling of the partition function~\cite{finn2016guided} or via a Laplace approximation~\cite{Levine2012ContinuousIO}. }

\changenew{To approximate the intractable integral in \eqref{eq:demo_boltzmann}, we sampled a set $\mathcal{X}$ of 1500 trajectories. We sampled costs according to \eqref{eq:demo_cost} given by random unit norm $\weight$s, then optimized them with an off-the-shelf trajectory optimizer. We used TrajOpt~\mbox{{\cite{trajopt}}}, which is based on sequential quadratic programming and uses convex-convex collision checking. This way, we obtain dynamically feasible trajectories that optimize for different features in varying proportions. While this sampling strategy cannot be justified theoretically, it works well in practice: the resulting optimized trajectories are a heuristic for sampling diverse and interesting trajectories in the environment. Future work will address this shortcoming by either providing theoretical guarantees or using importance sampling instead.}

For the second approximation to \eqref{eq:posterior_demo}, we discretized the space of $\weight\in\wspace$ and $\beta\in\mathcal{B}$ into sets $\wspace_D$ and $\mathcal{B}_D$, which leaves us with a finite, easy to compute posterior. 
For more practical details on specific discretization schemes, see Appendix \ref{app:practical_demos}.

    


Using the above discretization\footnote{In situations where the designer might want high fidelity inference over a large space of $\weight$ vectors, reasoning over a heavily discretized space would be more computationally expensive. However, longer offline computation is possible in our learning-from-demonstrations scenario as the inference happens offline, after providing the robot with human demonstrations. Alternatively, we could use Monte Carlo sampling approaches, similar to \cite{HadfieldMenell2017InverseRD,ramachandran2007bayesian}.}, we can now perform tractable inference \change{from demonstrations $\mathcal{D}$} to obtain a discrete posterior $b(\weight,\beta)$. \change{Algorithm \ref{alg:demos} summarizes the full procedure: given $\wspace_D, \mathcal{B}_D, \mathcal{X}$, and $\mathcal{D}$, our method iteratively updates the belief using \eqref{eq:demo_boltzmann} and \eqref{eq:posterior_demo}, resulting in the posterior $b(\weight,\beta)$. Lacking any a-priori information, we chose a uniform prior but our method will work with any prior.} We next present examples for what this posterior looks like in different scenarios.

\begin{algorithm}
\change{
    \caption{\change{Learning from Demonstrations (Offline)}}
    \label{alg:demos}
    \begin{algorithmic}
        \REQUIRE Discretized sets $\wspace_D, \mathcal{B}_D, \mathcal{X}$, set of demonstrations $\mathcal{D}$.
        \ENSURE Posterior belief $b(\weight,\beta)$ inferred from $\mathcal{D}$.
        \STATE $b(\weight, \beta) \gets Uniform(\weight, \beta).$
        \FOR{$\xtraj$ in $\mathcal{D}$}
            \FORALL{$\weight \in \wspace_D, \beta \in \mathcal{B}_D$}
            \STATE $P(\xtraj \mid \weight,\beta) = \frac{
            e^{-\beta\weight^T\Phi(\mathbf{x})}}{ \sum_{\bar\xtraj\in\mathcal{X}}e^{-\beta\weight^T\Phi(\bar\xtraj)}} \enspace\text{as per \eqref{eq:demo_boltzmann}}.$
            \STATE $b(\weight, \beta) \gets \frac{P(\xtraj \mid \weight,\beta)b(\weight, \beta)}{\sum_{\bar{\weight} \in \Theta, \bar{\beta} \in \mathcal{B}}P(\xtraj \mid \bar{\weight}, \bar{\beta})b(\bar{\weight}, \bar{\beta})} \enspace\text{as per \eqref{eq:posterior_demo}}.$
            \ENDFOR
        \ENDFOR
    \end{algorithmic}
}
\end{algorithm}


\begin{figure}[t!]
    \subfloat[(Left) Simulated perfect demonstration with the objective to keep the cup close to the table. (Right) Posterior belief resulted from this demonstration. Notice that a perfect demonstration leads to a high probability on the correct $\weight$ and high values for $\beta$.]{\label{spec}\includegraphics[width=\columnwidth]{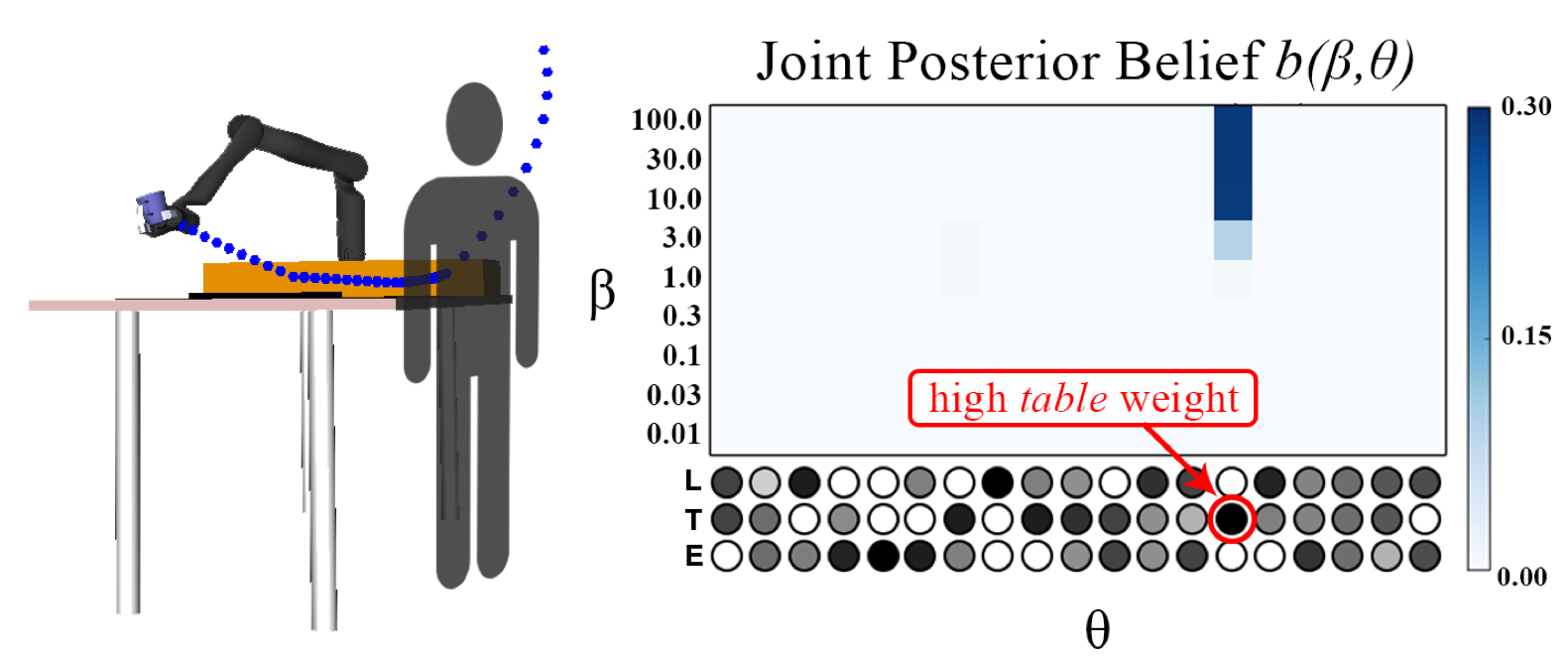}} \\
    \subfloat[(Left) Noisy human demonstration with the objective to keep the cup close to the table. (Right) Posterior belief resulted from this demonstration. Notice that a noisy but well-explained demonstration leads to a high probability on the correct $\weight$ and moderately high values for $\beta$. However, the noise in the demonstration significantly reduces the probability at the distributional peak.]{\label{noisy}\includegraphics[width=\columnwidth]{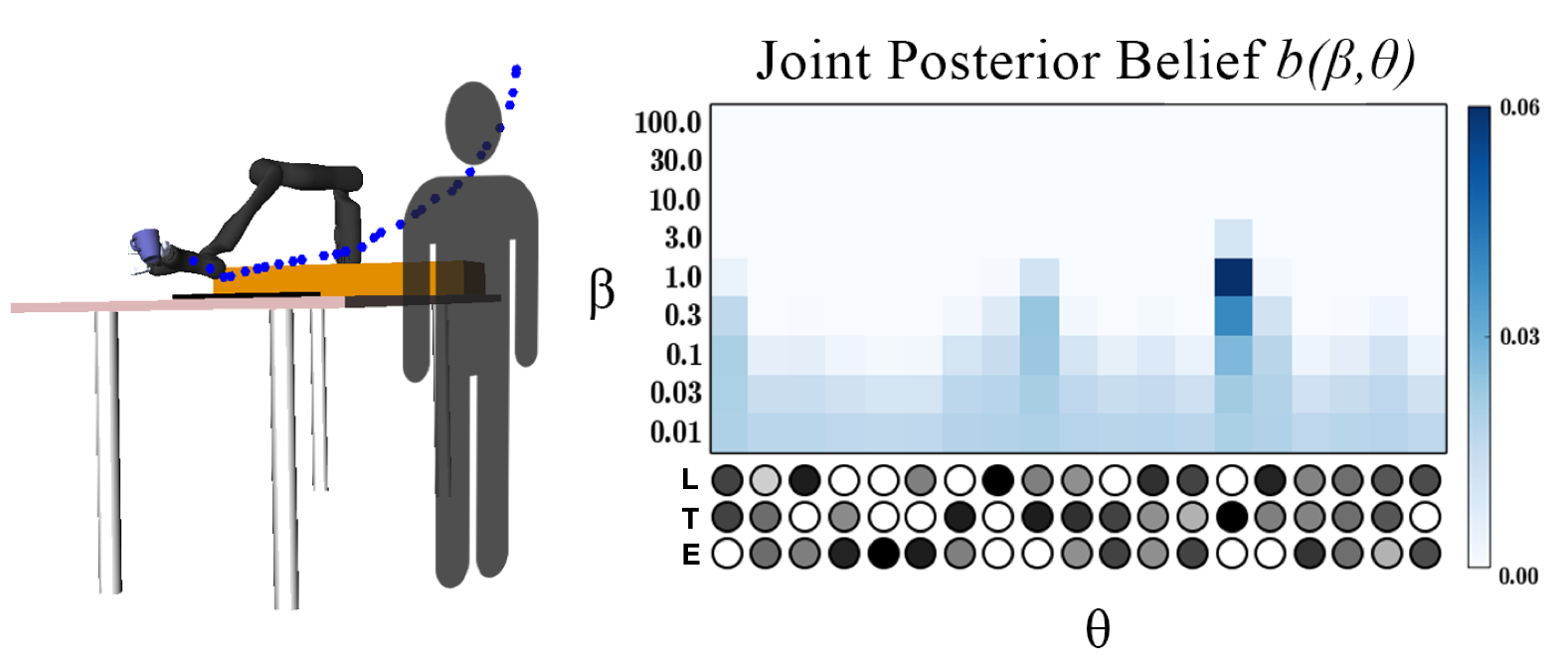}}
    \\
    \subfloat[(Left) Simulated perfect demonstration with the objective to keep the cup away from the human’s body. (Right) Posterior belief resulted from this demonstration. Notice that, since this demonstration is poorly explained (the robot is not reasoning about distance from the human), the posterior belief is spread out approximately uniformly over all $\weight$s and the lowest $\beta$ values. This indicates that the robot cannot tell what the demonstration was intended for.]{\label{misspec}\includegraphics[width=\columnwidth]{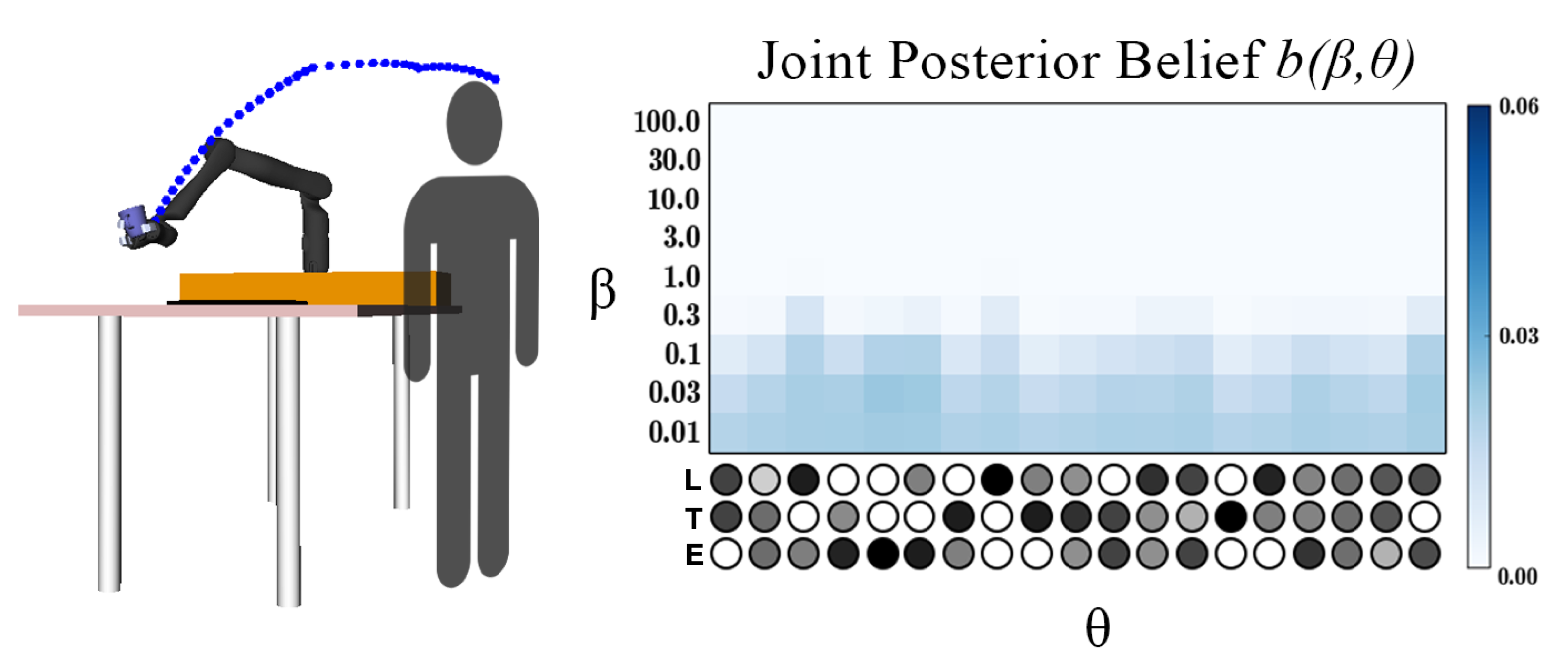}}
    \caption{Three examples of demonstrations and the inferred posterior belief after each one of them. The robot infers the right $\weight=[0,1,0]$ from the two well-explained demonstrations, but, unlike the perfect simulated demonstration in \ref{spec}, the noisy one in \ref{noisy} cannot reach the highest $\beta$ and has as overall more spread-out probability distribution with a lower peak value. Lastly, the perfect simulated demonstration that is poorly explained in \ref{misspec} results in a posterior that is spread-out over all $\weight$s and the lowest $\beta$s , consistent with the robot not being able to tell what the human’s objective was.}
    \label{fig:simulated_demo}
\end{figure}


\subsection{Examples}

To provide intuition for how situational confidence can indicate when a robot's hypothesis space is misspecified, we illustrate some examples with a robot manipulator learning from a human demonstrator. \change{These examples help prepare the setup we will present in our actual experiments in Section \ref{sec:case_study}.}

The robot manipulator is performing a household task of moving cups from a shelf onto the kitchen table. The robot needs to learn from the person's demonstrations how to best perform this task. For this purpose, the person physically guides the robot through one or a few demonstrations of moving the cup down to the table, from which the robot infers the hidden objective function.

In the\change{se} examples\strike{ to follow}, the robot's hypothesis space includes three features: efficiency \change{(E)} \strike{in the form of} \change{as} sum of squared velocities over the trajectory, keeping the cup close to the table \change{(T)}, and keeping the cup away from the laptop \change{(L)} depicted in black.
    
Formally, we can represent these three feature mappings as:
    \begin{equation}
        \trajfeat(\xtraj) =
        \begin{bmatrix}
        \sum^T_{i=1} ((\state^i-\state^{i-1})/\Delta t)^2 \\
        \sum^T_{i=0} ||\state^i - \state_\text{table}||_2 \\
        \sum^T_{i=0} \max\{0, L - ||\state^i - \state_\text{laptop}||_2\} 
        \end{bmatrix}
    \label{eq:features}
    \end{equation}
    
    where $L$ is the radius of a \change{penalty }sphere around the laptop\strike{ where penalties apply}, $\Delta t$ is the discrete timestep between the states in the trajectory, and the corresponding feature weight vector is $\weight \in \mathbb{R}^3$.

\begin{figure*}[t!]
    \centering
    \subfloat[In the true graphical model, $\control_H$ is an observation of $\weight$ and the situational confidence $\beta$.]{\label{subfig:model0}\includegraphics[width=0.4\columnwidth]{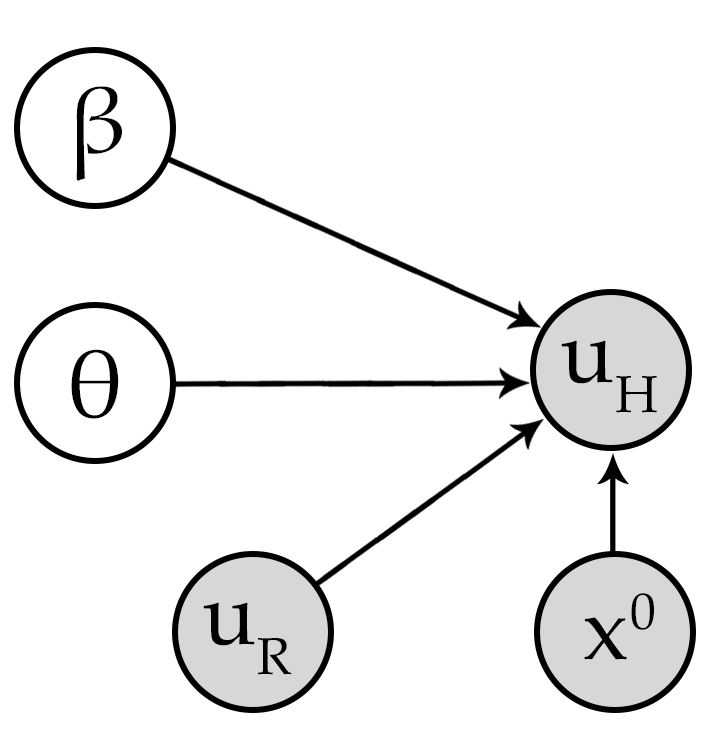}}
    \hspace{8mm}
    \subfloat[We use the proxy variable $\Phi$ to first estimate $\beta$ efficiently.]{\label{subfig:model1}\includegraphics[width=0.5\columnwidth]{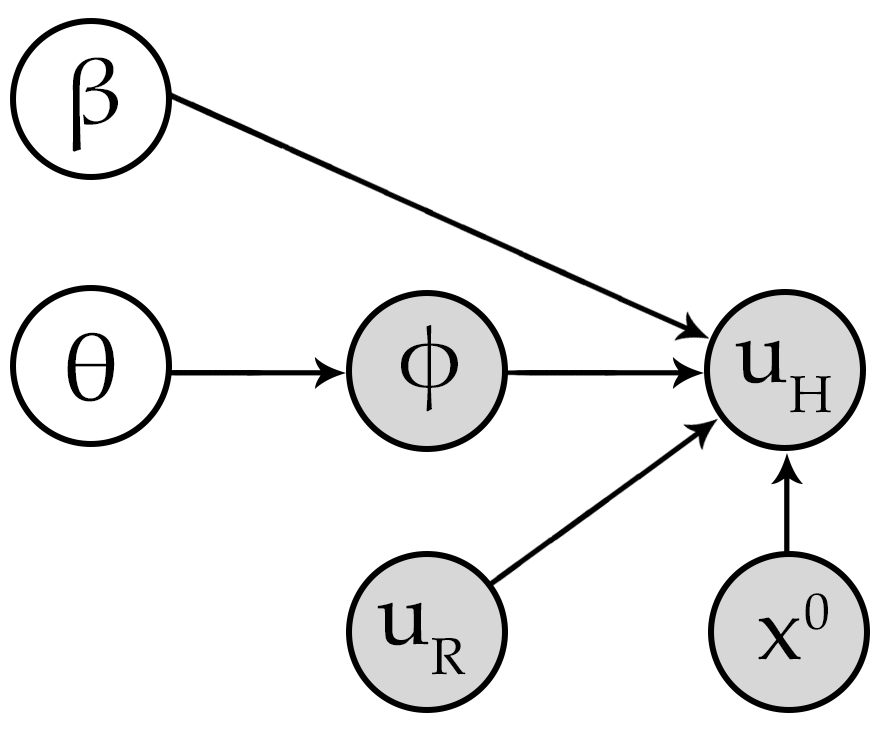}}
    \hspace{5mm}
    \subfloat[We interpret the estimate $\hat\beta$ as an indirect observation of the unobserved $E$, which we then use for the $\weight$ estimate.]{\label{subfig:model2}\includegraphics[width=0.7\columnwidth]{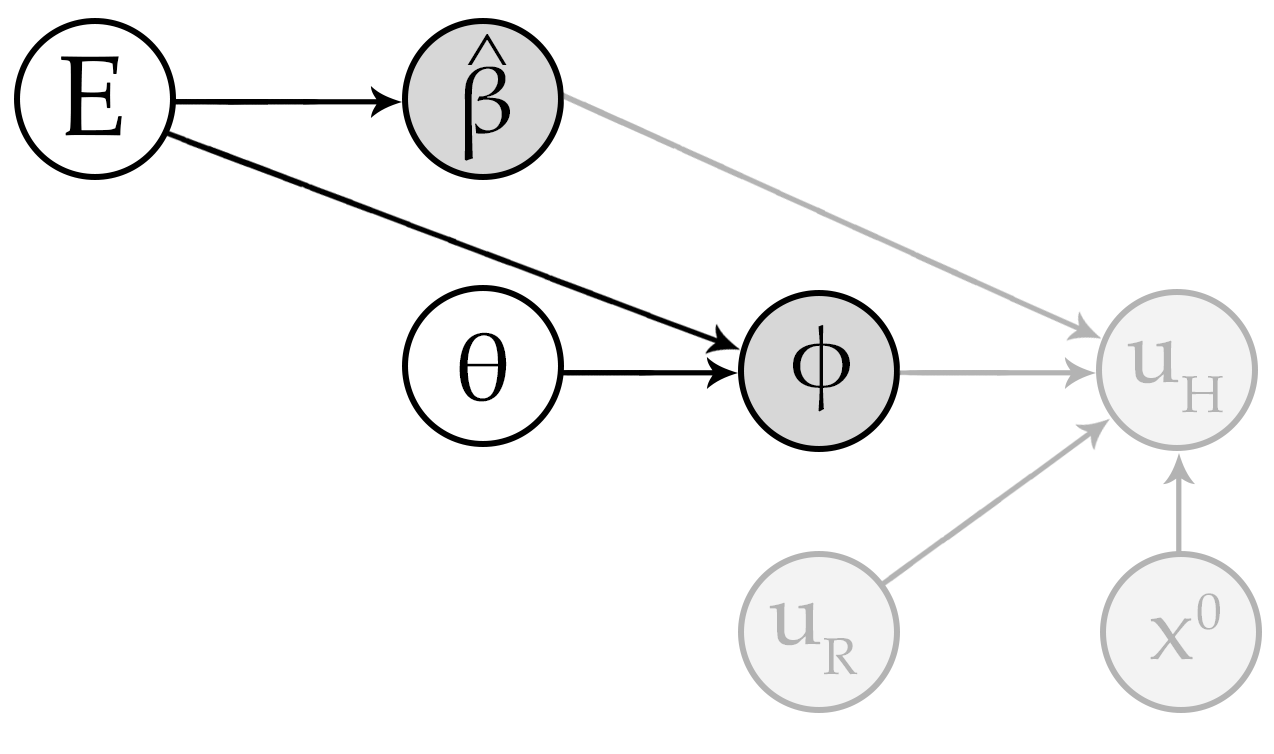}}
    \caption{Graphical model formulation (a) and modifications to it ((b) and (c)) for real-time tractability.}    \label{fig:graphical_models}
\end{figure*}

Fig. \ref{fig:simulated_demo} demonstrates how the feature weight $\theta$ and the situational confidence $\beta$ are affected for well-explained, noisy, and poorly-explained simulated human demonstration. The posterior belief is shown for the combination of discrete parameters $\weight$ and $\beta$. Higher $\beta$ values indicate higher situational confidence. The three circles under each column represent the $\weight$ vector for that column, with the components being the efficiency, distance from the table, and distance from the laptop features. A larger feature weight is indicated by a darker colored circle, while a white color indicates zero weight.

First, in \ref{spec}, we consider the case where the demonstration is a perfectly optimal trajectory produced by TrajOpt~\cite{trajopt}. This serves as a sanity check for when the human and the robot have the same hypothesis space and the demonstration is perfect. The optimal demonstration was produced by finding a trajectory that moves the cup from the start configuration to the end while minimizing the distance between the cup and the table. Notice that, with a perfect demonstration, the posterior distribution places the most probability mass on the $\weight$ that indicates high penalties for staying away from the table but no penalties for lack of efficiency or closeness to laptop. Moreover, the posterior also reveals that the most likely $\weight$ also corresponds \textit{with the highest available confidence $\beta$}.

Next, \change{in \ref{noisy}} we recorded a real human demonstration of the same cup-to-table behavior. The nature of demonstrations both on hardware and from real people introduce noise into the demonstration, making it potentially suboptimal with respect to the robot's model. However, in this case the human and the robot still share the same hypothesis space (i.e. the robot and the human both know about the the efficiency, table, and laptop features). Here, we study how the noise in the demonstration affects the robot's inference. Notice that even with an imperfect demonstration, the robot is able to identify the correct $\weight$ parameter, but now with a lower confidence $\beta$.
    
Lastly, we consider the example where the demonstration is optimal but the robot does not have a rich enough hypothesis space to explain it. The robot reasons about the same three features, but now the demonstration was produced by optimizing for an additional feature that is outside its hypothesis space: keeping the cup away from the human's body. We observe that the probability distribution \change{in \ref{misspec}} is spread over all the $\weight$ values in the space, with the highest values on low $\beta$s. This example shows how, in the case of poorly-explained input, the robot's inference is unsure which objective the human had in mind, and assigns low situational confidence to the given input.

These illustrative examples give us valuable insight into how the \change{($\weight$, $\beta$)-belief} changes depending on how well-explained the input is. For perfectly explained demonstrations, the inference identifies the correct $\weight$ with high posterior probability. As the input becomes more poorly-explained, the robot loses confidence in all $\weight$s, assigning approximately uniformly spread-out probability on the lowest situational confidence values $\beta$.

\section{Algorithmic Approach: Corrections}
\label{sec:corrections}

\subsection{Formulation}

We consider the setting in which human input is provided in the form of physical interventions during the robot's task execution. Fig. \ref{fig:demos_and_corrections} (right) \strike{illustrates} \change{is} an example of such a correction. 
The human may provide a correction \strike{seeking }to improve some aspect of the task execution that is not represented in the robot's \strike{modeled feature} \change{objective} space.
When the robot receives \strike{human }input, it should be able to reason about its situational confidence in light of the correction and replan its trajectory accordingly for the \strike{remainder} \change{rest} of the task execution or until a new correction \strike{is received} \change{happens}.
Thus, the robot must have access to an inference algorithm that can run in real time.
In this section, we will present an online version of our situational confidence framework.


In the physical corrections setting, the robot starts with an initial guess of the parameter~$\weight$ and uses a trajectory optimization scheme to compute a motion plan seeking to minimize the associated cost~$\cost_\weight$.
The robot performs the task at hand by applying controls~$\utraj_R$ via an impedance controller in order to track the computed trajectory~$\xtraj$. 

At any timestep $t$ during the trajectory execution, the human may physically interact with the robot, inducing a joint torque $\control_H^t$. When this happens, the robot can use the human input to update its estimated $\weight$ parameter, and thereby the corresponding objective $\cost_\weight$.
Given the new adapted objective, the robot replans an optimized trajectory $\xtraj$ and tracks it until the next human input is sensed or until the task is completed.

Following \cite{bajcsy2017phri}, the robot's representation of the task assumes that the human does not explicitly care about the robot's control effort, but only about features of the state trajectory.
In addition, the human is assumed to have a preference for minimizing her own control effort.
This captures the human's incentive to have the robot perform the task autonomously, providing only minimal input to guide the robot towards the correct behavior when necessary.
Encompassing these assumptions, the cost \eqref{eq:linear_cost} takes the form:
\begin{equation}
\label{eq:phri_cost}
    \cost_{\weight}(\xtraj, \control_H^t) = ~\weight^T\trajfeat(\xtraj) + \lambda\|\control_H^t\|^2.
\end{equation}
To approximately compute the trajectory resulting from the human's input, we follow the approach in \cite{bajcsy2017phri} and introduce the notion of a \textit{deformed trajectory} $\xtraj_D$. This trajectory constitutes the robot's estimate of the human's desired trajectory given her applied torque $\control_H^t$.
Given the robot's default trajectory $\xtraj_R:=\xtraj(\cdot;\state^0,\utraj_R,\mathbf{0})$ and having observed the instantaneous human intervention $\control_H^t$, we compute $\xtraj_D$ by deforming the robot's default trajectory in the direction of $\control_H^t$:
\begin{equation}
    \begin{aligned}
        \xtraj_D = \xtraj_R + \mu A^{-1}\tilde\utraj_H
    \label{eq:deformation}
    \end{aligned}
    \enspace,
\end{equation}
where $\mu>0$ scales the magnitude of the deformation, $A \change{\in \mathbb{R}^{n(T+1) \times n(T+1)}}$ defines a norm on the Hilbert space of trajectories\footnote{%
    We used a norm $A$ based on acceleration, consistent with \cite{bajcsy2017phri}, but other norm choices are possible as well.%
} and dictates the deformation shape \cite{deformation}, and $\tilde\utraj_H\change{\in \mathbb{R}^{n(T+1)}}$ is $\control_H^t$ at \change{indices $nt$ through $n(t+1)$} and 0 otherwise.  
%
The human is
therefore modeled by \eqref{eq:phri_cost} 
as trading off between inducing a good trajectory $\xtraj_D$ with respect to~$\weight$, and minimizing her effort.
    
Equipped with this cost function, we \strike{now }need the robot to reason about the reliability of its objective space given new inputs in the form of corrections.
In contrast with our analysis in Section \ref{sec:demonstrations}, here the person does not give full demonstrations $\xtraj$, but instead offers corrections $\control_H^t$ based on the robot's default trajectory $\xtraj_R$.
Applying \eqref{eq:boltzmann} to this setting, we have:
\begin{equation}
\label{eq:boltzmann_phri}
    P(\control_H^t \mid \state^0, \utraj_R ; \weight, \beta) = \frac{e^{-\beta ({\weight}^\top\Phi(\xtraj_D) + \lambda\|\control_H^t\|^2)}}{\int e^{-\beta ({\weight}^\top\Phi(\bar{\xtraj}_D) + \lambda\|\bar{\control}\|^2)}d\bar{\control}}\enspace,
\end{equation}
where $\xtraj_D$ and $\bar{\xtraj}_D$ are given by \eqref{eq:deformation} applied to their respective controls $\control_H^t$ and $\bar{\control}$.

Ideally, with this model of human actions, illustrated in Fig. \ref{subfig:model0}, we would perform inference over both the situational confidence $\beta$ and the modeled parameters $\weight$ by maintaining a joint Bayesian belief $b'(\weight, \beta)$. Analogously to the demonstrations case, our probability distribution over $\weight$ would automatically adjust for well-explained corrections, whereas for poorly-explained ones the robot's posterior would not deviate significantly form its prior on $\weight$.
Unfortunately, this Bayesian update is not generally feasible in real time, given the continuous and possibly high-dimensional nature of the parameter space $\wspace$.
Even in simple scenarios with a small number of continuous features, discretizing $\wspace$ as we did in the demonstrations case would generally yield an overly slow inference, making the method impractical for use in the real-time collaborative scenarios that we are interested in here. \change{Thus, to evaluate the benefits of estimating $\beta$ we need to derive an online method that goes beyond simple discretization.}

\subsection{Approximation} 

To alleviate the computational challenge of performing joint inference over $\beta$ and $\weight$, we introduce a structural assumption that will enable us to approximately decouple the two inference problems.

\subsubsection{Estimating $\beta$} 

To estimate $\beta$ without dependence on $\weight$, we will assume that in order to decide what correction to provide, the human will first choose the desired features $\Phi$ of the resulting trajectory $\xtraj_D$ and then select an input $\control_H^t$ that will obtain these features (Fig. \ref{subfig:model1}).

Based on the observed human input $\control_H^t$ and the trajectory features of the deformed trajectory $\Phi(\xtraj_D)$, the robot can obtain an estimate of \change{$\beta$} by considering how efficient the human's input was for the features achieved. Letting $\uset_\Phi$ be the set of inputs that achieve the same observed features $\Phi_D:=\Phi(\xtraj_D)$, the Boltzmann decision model gives
\begin{align}
\label{eq:boltzmann_just_for_u}
    P(\control_H^t \mid \state^0, \utraj_R ; \Phi_D, \beta) &= \frac{e^{-\beta ({\weight}^\top\Phi_D + \lambda\|\control_H^t\|^2)}}{\int_{\uset_\Phi} e^{-\beta ({\weight}^\top\Phi(\bar{\xtraj}_D) + \lambda\|\bar{\control}\|^2)}d\bar{\control}}\notag\\
    &=\frac{e^{-\beta \lambda\|\control_H^t\|^2}}{\int_{\uset_\Phi} e^{-\beta \lambda\|\bar{\control}\|^2}d\bar{\control}}
    \enspace,
\end{align}
since the term ${\weight}^\top\Phi(\bar\xtraj_D)$ is constant for all $\bar\control\in\uset_\Phi$ and equal to the term ${\weight}^\top\Phi_D$ in the numerator.

Using \eqref{eq:boltzmann_just_for_u}, the robot can obtain an estimate of $\beta$ by considering how efficient the human's correction was for the features achieved---if the input seems highly inefficient, this is indicative that the features modeled by the robot may not accurately capture the human's preference.
\strike{Depending on whether a prior distribution over $\beta$ is available, two appropriate estimates for $\beta$ may be the maximum likelihood estimator (MLE) or the maximum \emph{a posteriori} (MAP) estimator.}

It is useful to approximate the integral over the constrained set $\uset_\Phi\subset\uset$ by an integral over the entire set of possible inputs $\uset$, introducing a \strike{Lagrangian} \change{penalty} term in the exponent that results in a soft indicator function for $\bar\control\in\uset_\Phi$:
\begin{equation}
\label{eq:bayesnetb_model}
	 P(\control_H^t \mid \state^0, \utraj_R ; \Phi_D, \beta)
	 \approx
	 \frac{e^{-\beta \lambda\|\control_H^t\|^2}}{\int_\uset e^{-\beta( \lambda\|\bar{\control}\|^2+ \change{\kappa}\|\trajfeat(\bar{\xtraj}_D) - \Phi_D\|^2 )}d\bar{\control}}
	\enspace.
\end{equation}
Note that for an arbitrarily large \strike{Lagrange multiplier $\nu$} \change{$\kappa$ }
there is an arbitrarily small probability assigned to $\uset\setminus\uset_\Phi$ in the integral. It is now possible to apply the Laplace approximation to the unconstrained integral (see Appendix \ref{app:laplace} for details), yielding:
\begin{align}
\label{eq:laplace}
	P(\control_H^t &\mid \state^0, \utraj_R ; \Phi_D, \beta) \approx \notag
	\\&  \frac{e^{-\beta\lambda  \| \control_H^{\change{t}} \|^2}}{ e^{-\beta(\lambda \|{\control_H^*}\|^2 + \change{\kappa}\|\trajfeat(\xtraj_D^*) - \trajfeat_D\|^2)}}\sqrt{\frac{\beta^k|H_{\control^*_H}|}{2\pi^k}}
    \enspace,
\end{align}
where $k$ is the action space dimensionality and $H_{\control_H^*}$ is the Hessian of the \strike{cost function} \change{exponent in the denominator of \eqref{eq:bayesnetb_model}} around $\control_H^*$. We obtain the optimal action $\control_H^*$ by solving the constrained optimization problem \change{(see Appendix \ref{app:practical_corrections})}: 
\begin{equation}
\begin{aligned}
& \underset{\change{\tilde\control_H}}{\text{minimize}}
& & \|\change{\tilde\control_H}\|^2 \\
& \text{subject to}
& & \trajfeat(\xtraj+\mu A^{-1}\tilde\utraj_H) - \trajfeat_D = 0
\enspace.
\end{aligned}
\label{opt:optimal_uH}
\end{equation}
In other words, the resulting $\control_H^*$ is the minimal norm $\tilde\control_H$ the human could have taken, constrained to lie in $\uset_\trajfeat$.
\change{As such, the second norm in the denominator's exponent is 0, and} the final conditional probability becomes:
\begin{equation}
	P(\control_H^t \mid \state^0, \utraj_R ; \Phi_D, \beta) = e^{-\beta\lambda(\|\control_H^t\|^2-\|\control_H^*\|^2 )}\sqrt{\frac{\beta^k|H_{u^*_H}|}{2\pi^k}}
	\enspace.
\label{eq:laplace2}
\end{equation}
We derive below the maximum likelihood estimator (MLE), noting that a maximum \emph{a posteriori} (MAP) estimator is often appropriate given a certain prior on $\beta$.
\begin{equation}
\begin{aligned}
	\hat{\beta} =& \arg\max_{\beta} \{\log(P(\control_H^t \mid \state^0, \utraj_R ; \Phi_D, \beta)\} \\
	=& \arg\max_{\beta} \{-\beta\lambda(\|\control_H^{\change{t}}\|^2-\|\control_H^*\|^2 ) + \log(\sqrt{\frac{\beta^k|H_{\control^*_H}|}{2\pi^k}})\}.
\end{aligned}
\end{equation}
Applying the first-order condition and setting the derivative to zero yields the maximizer:
\begin{equation}
	\hat{\beta} = \frac{k}{2\lambda(\|\control_H^{\change{t}}\|^2-\|\control_H^*\|^2)}\enspace.
\label{eq:beta_map}
\end{equation}

\begin{figure*}[t!]
\centering
  \centering
  \includegraphics[scale=.47]{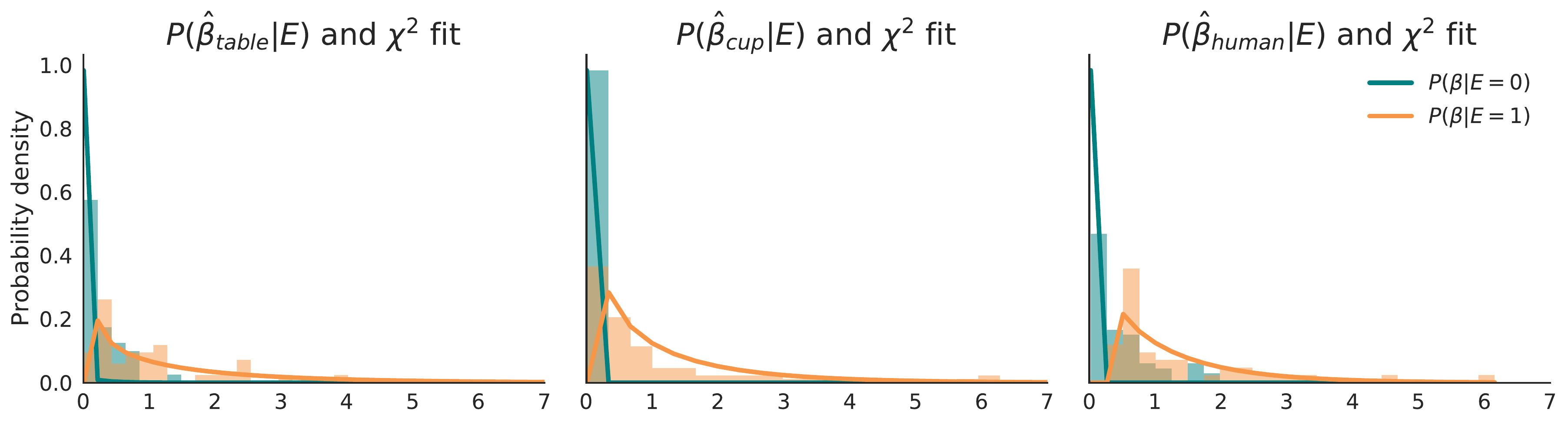}
  \caption{Empirical estimates for $P(\hat{\beta} \mid E)$ and their corresponding chi-squared ($\chi^2$) fits.}
  \label{fig:histograms}
\end{figure*}

The estimator\change{\footnote{\change{Note that $\hat\beta$ is non-negative, since $\control_H^*$ is the minimal-norm $\tilde\control_H$ that satisfies the constraint, so the difference in the denominator of \eqref{eq:beta_map} is positive.}}} above yields a high value when the difference between $\control_H^{\change{t}}$ and $\control_H^*$ is small, i.e. the person's correction achieves the induced features $\trajfeat(\xtraj_D)$ efficiently. For instance, if $\xtraj_D$ brings the robot closer to the table, and $\control_H^{\change{t}}$ pushes the robot straight towards the table, $\control_H^{\change{t}}$ is an efficient way to induce those new feature values. However, when there is a much more efficient alternative (e.g. when the person pushes mostly sideways rather than straight towards the table), $\hat{\beta}$ will be small. Efficient ways to induce the feature values will suggest well-explained inputs, inefficient ones will suggest poorly-explained corrections.

\subsubsection{Estimating ${\weight}$}

\change{To} tractably estimate $\weight$ building on the $\beta$ estimate, we introduce an auxiliary binary variable ${E \in \{0,1\}}$ indicating whether the human's intervention can be well \textit{explained} by the robot's modeled cost features.
We will perform offline training with ground-truth access to this variable in order to \strike{learn} learn its relation to the robot's estimate $\hat\beta$.

When $E=1$, the human's desired modification of the robot's behavior can be well explained by \emph{some} vector ${\weight \in \wspace}$, which will lead the intervention to appear less noisy to the robot (i.e. $\beta$ is large).
As a result, the correction~$\control_H^t$ is likely to be efficient for the cost encoded by this $\weight$.
Conversely, when $E=0$, the intervention appears noisy (i.e. $\beta$ is small), and the human's correction cannot be well explained by any of the cost features modeled by the robot.

The graphical model depicted in Fig. \ref{subfig:model2} relates the induced feature values $\trajfeat_D$ to $\weight$ as a function of the $E$. When $E=1$, the induced features will tend to have low cost with respect to $\weight$; when $E=0$, the induced features \emph{do not depend on} $\weight$, and we model them as Gaussian noise centered around the feature values of the robot's currently planned trajectory $\xtraj_R$.
\begin{equation}
    P(\trajfeat_D \mid \weight, E) =
    \begin{cases}
        \displaystyle\frac{e^{-\weight^\top\trajfeat_D}}{\int e^{-\weight^{\top}\trajfeat(\tilde{\xtraj}_D)} d\tilde{\xtraj}_D} ,& E=1 \\[4mm] \left(\frac{\nu}{\pi}\right)^\frac{k}{2} e^{-\nu ||\trajfeat_D-\trajfeat(\xtraj_R)||^2} ,& E=0
    \end{cases}
\label{eq:bayesnetc_model}
\end{equation}
with the constant in the $E=0$ case corresponding to the normalization term of the normal distribution.

In addition, this graphical model relates the $\hat{\beta}$ resulting from the model in Fig. \ref{subfig:model1} to $E$ by a $P(\hat{\beta} \mid E)$. We fit this distribution from controlled user interaction samples where we have ground-truth knowledge of $E$\change{\footnote{\change{Since we tell users what to optimize for, we know whether the human's input is well-explained with respect to the robot's hypothesis space or not.}}}.
For each sample interaction, we compute $\hat{\beta}$ (for example, using \eqref{eq:beta_map} if using MLE) and label it with the corresponding binary $E$ value.
We fit a chi-squared distribution to these samples to obtain the probability distributions for $P(\hat{\beta} \mid E=0)$ and $P(\hat{\beta} \mid E=1)$. The resulting distributions are shown in Fig. \ref{fig:histograms}\change{\footnote{\change{Because users tend to accidentally correct more than one feature, we perform $\beta$-inference separately for each feature. This requires more overall computation (although still linear in the number of features, and can be parallelized) and a separate $P(\hat{\beta} \mid E)$ estimate for each feature.}}}.


Using the model in Fig.~\ref{subfig:model2} with the learned distribution $P(\hat\beta\mid E)$, we can infer a~$\weight$ estimate in real time whenever a physical correction from the human is measured.
We do this tractably by interpreting the estimate~$\hat\beta$ obtained from~\eqref{eq:beta_map} as an indirect observation of the unknown variable $E$.
We combine the empirically characterized likelihood model $P(\hat\beta\mid E)$ with an initial uniform prior $P(E)$ to maintain a Bayesian posterior on $E$ based on the evidence $\hat\beta$ constructed from human observations at deployment time, $P(E \mid \hat\beta) \propto P(\hat\beta \mid E) P(E)$.


Further, since we wish to obtain a posterior estimate of the human's objective $\weight$, we use the model from Fig. \ref{subfig:model2} to obtain the posterior probability measure
\begin{equation}\label{eq:theta_from_beta_hat}
    P(\weight \mid \trajfeat_D,\hat\beta) \propto \sum_{E \in \{0,1\} } P\big(\trajfeat_D \mid \weight, E\big) P(E \mid \hat\beta) P(\weight)
\enspace.
\end{equation}

Following \cite{bajcsy2017phri}, \change{we note that we can approximate the partition function in the human's policy \eqref{eq:bayesnetc_model} by employing the Laplace approximation. Taking a second-order Taylor series expansion of the exponent's objective about $\xtraj_R$, the robot's current best guess at the optimal trajectory, we obtain a Gaussian integral that can be evaluated in closed form}
\strike{we approximate the partition function in the human's policy \mbox{\eqref{eq:bayesnetc_model}} by the exponentiated cost of the robot's original trajectory, thereby obtaining}
\begin{equation}
P(\trajfeat_D \mid \weight, E=1) \change{\approx} 
	e^{-\weight^\top\big(\trajfeat_D-\trajfeat(\xtraj_R)\big)}\enspace. 
\label{eq:bayesnetc_with_constants}
\end{equation}

\begin{figure*}[t!]
    \centering
    \subfloat[(Left) The human applies well-explained corrections to keep the cup close to the table. Learning with fixed $\beta$ leads to a correct trajectory that satisfies the human's preference. (Right) As the person corrects the robot by pushing down on it, the learning algorithm gradually updates its weight on the feature modeling distance to table.]{\label{nob_spec}\includegraphics[width=0.9\columnwidth]{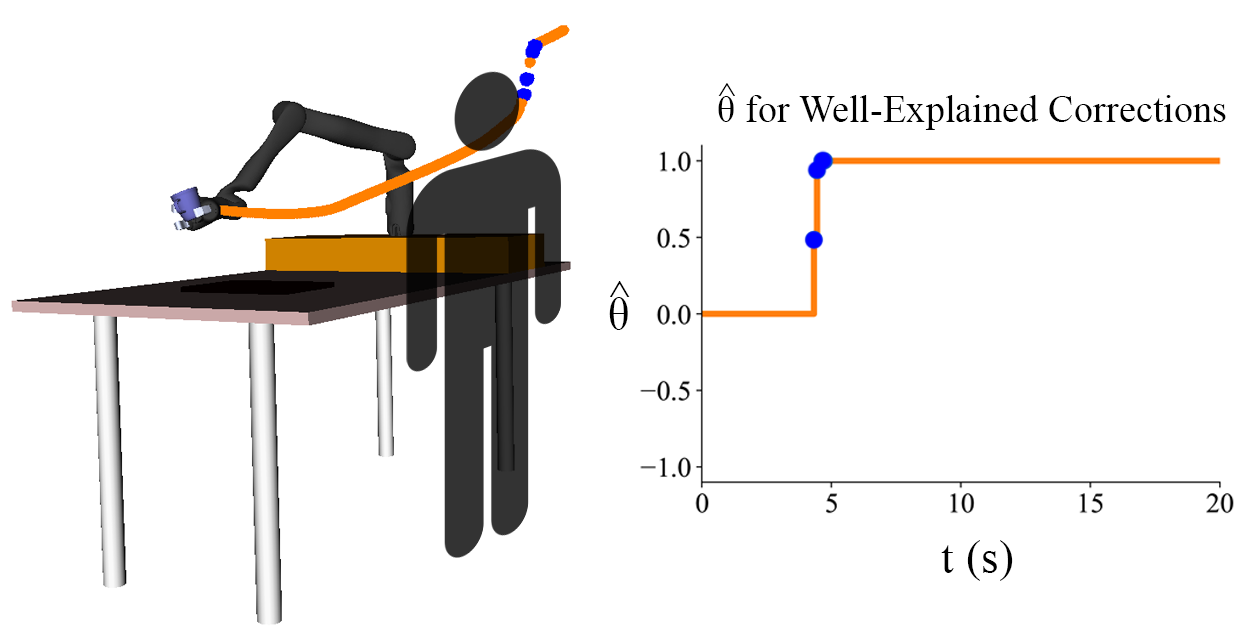}}
    \hspace{2mm}
    \subfloat[(Left) The human applies poorly-explained corrections to keep the cup upright. Learning with fixed $\beta$ leads to a oscillatory and noisy trajectory. (Right) Here the learning algorithm incorrectly updates the weight on the distance to table feature, leading to unintended learning.]{\label{nob_misspec}\includegraphics[width=0.9\columnwidth]{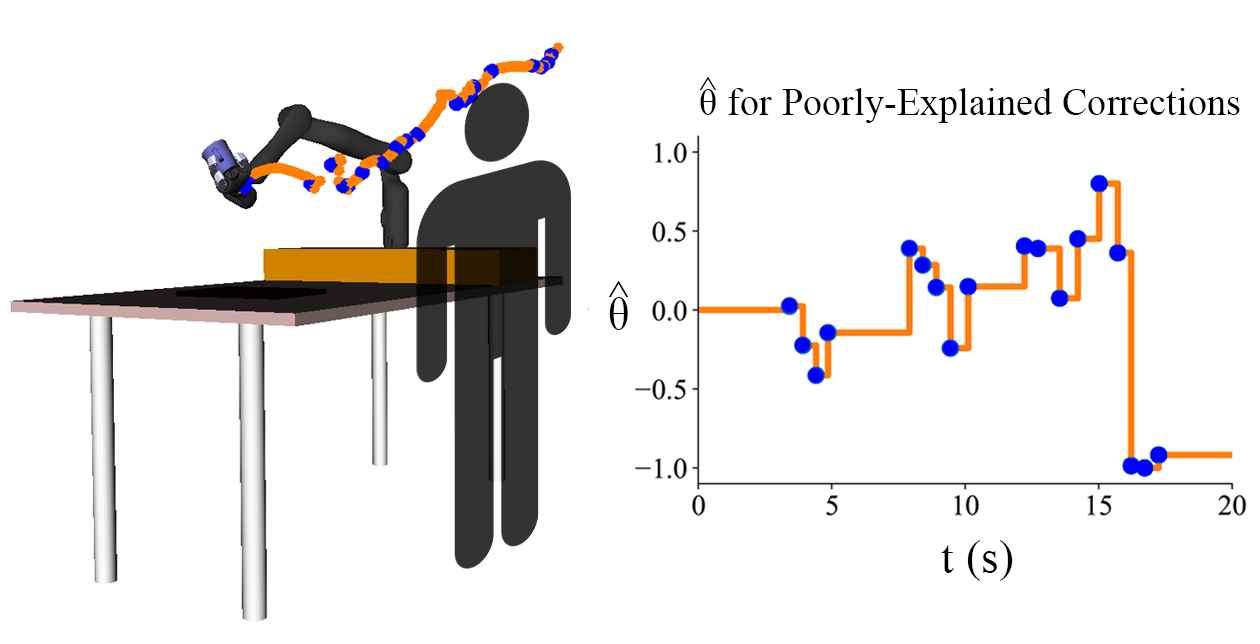}}
    \\
    \subfloat[(Left) The human applies well-explained corrections to keep the cup close to the table. Learning with estimated $\beta$ leads to a correct trajectory that satisfies the human's preference. (Right) As the person corrects the robot by pushing down on it, the learning algorithm infers high $\hat\beta$ and gradually updates its weight on the feature modeling distance to table.]{\label{b_spec}\includegraphics[width=0.9\columnwidth]{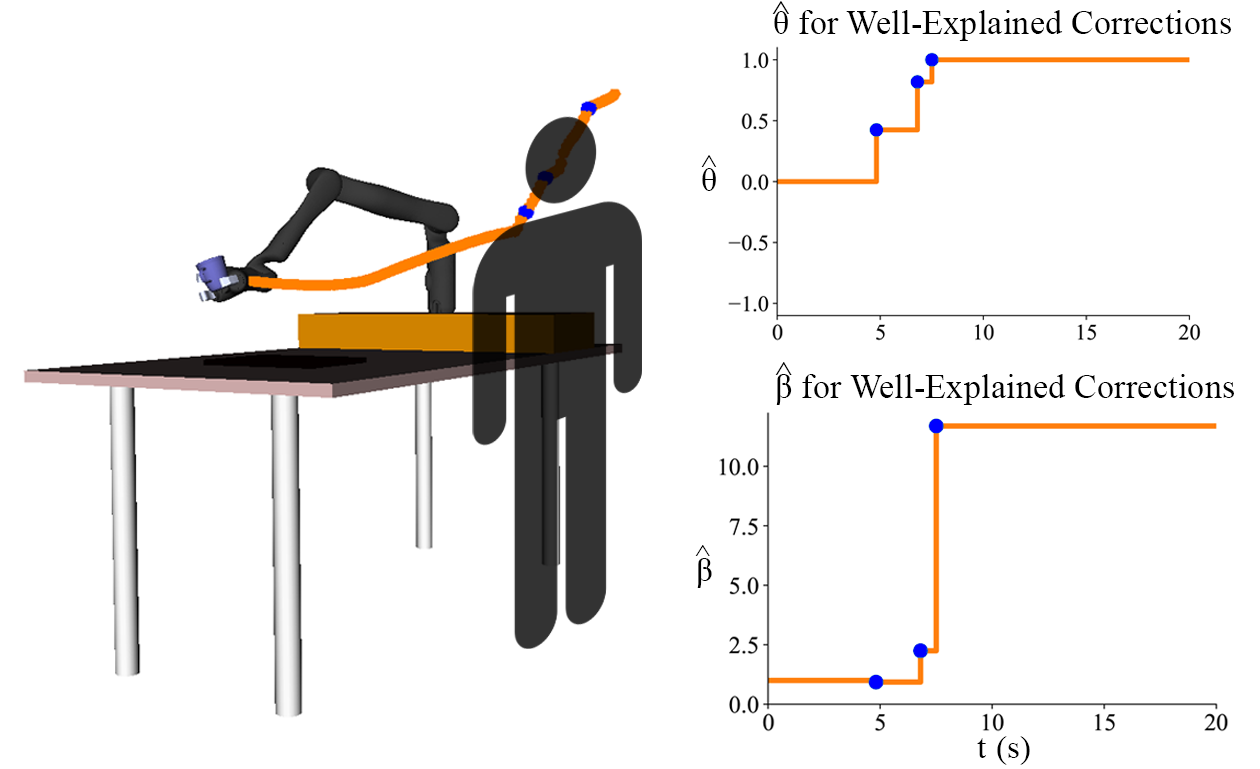}}
    \hspace{2mm}
    \subfloat[(Left) The human applies poorly-explained corrections to keep the cup upright. Learning with estimated $\beta$ leads to a smooth trajectory where the robot is robust to poorly-explained inputs.  (Right) Here the learning algorithm infers low $\hat\beta$ and correctly avoids unintended learning for the distance to table feature.]{\label{b_misspec}\includegraphics[width=0.9\columnwidth]{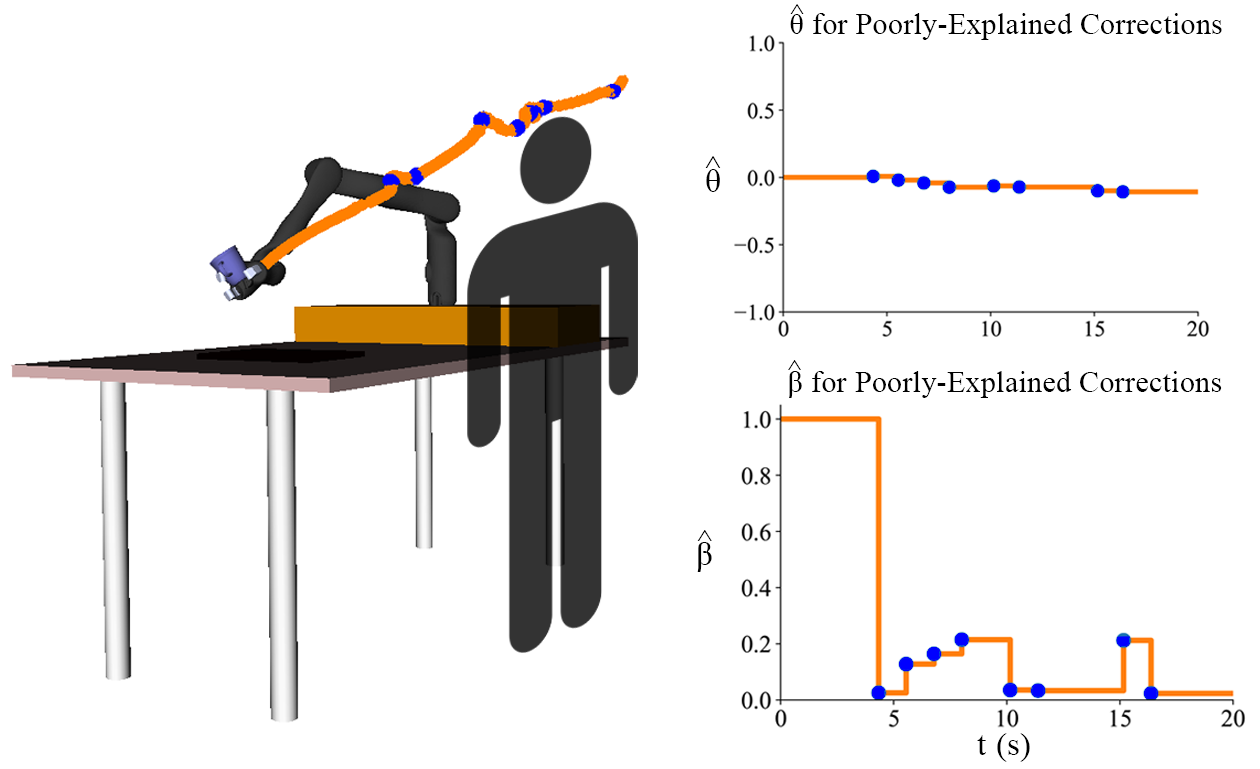}}
    \caption{Examples of physical corrections (interaction points shown in blue) and the resulting behavior for the fixed $\beta$ method (top) and estimated $\beta$ method (bottom). When the corrections are well explained, both methods learn the correct weight $\hat\weight=1.0$. In the case of poorly-explained corrections, our method infers low $\hat\beta$ and manages to reduce unintended learning, whereas the fixed $\beta$ method produces incorrect oscillatory behavior.}
    \label{fig:phri_examples}
\end{figure*}

We also consider a Gaussian prior distribution of $\weight$ around the robot's current estimate $\hat\weight$:
\begin{equation}\label{eq:prior}
P(\weight) = \frac{1}{(2\pi\alpha)^\frac{k}{2}} e^{-\frac{1}{2\alpha} ||\weight-\hat{\weight}||^2}
\enspace,
\end{equation}
\change{where $\alpha \geq 0$ determines the variance of the Gaussian.}

\change{To obtain an update rule for the $\weight$ parameter, we can simply plug \eqref{eq:bayesnetc_model}, \eqref{eq:bayesnetc_with_constants}, and \eqref{eq:prior} into \eqref{eq:theta_from_beta_hat}.} \change{For legibility, let's denote $\Gamma(\trajfeat_D, E=i) = P(E=i \mid \hat\beta)P\big(\trajfeat_D\mid\weight,E=i\big)$, for $i \in \{0, 1\}$. Then,} the maximum-a-posteriori estimate of the human's objective $\weight$ is the solution maximizer of
\begin{equation}
\begin{split}\label{eq:posterior_theta}
P(\weight) \Big[& \Gamma(\trajfeat_D, E=1) + \Gamma(\trajfeat_D, E=0) \Big] 
    \\ =\;&  \frac{1}{(2\pi\alpha)^\frac{k}{2}}
e^{-\frac{1}{2\alpha}||\weight-\hat{\weight}||^2}
   \Big[ P(E=1 \mid \hat\beta) e^{-\weight^\top\big(\trajfeat_D-\trajfeat(\xtraj_R)\big)}
   \\ &+ P(E=0 \mid \hat\beta) \left(\frac{\nu}{\pi}\right)^\frac{k}{2} e^{-\nu||\trajfeat_D-\trajfeat(\xtraj_R)||^2}\Big]
\enspace.
\end{split}
\end{equation}

Differentiating \eqref{eq:posterior_theta} with respect to $\weight$ and equating to 0 gives the maximum-a-posteriori update rule 
\change{
\begin{gather}\label{eq:theta_update}
 \hat\weight' = \hat\weight - \alpha \frac{\Gamma(\trajfeat_D, E=1)}
 {\Gamma(\trajfeat_D, E=1) + \Gamma(\trajfeat_D, E=0)}
 \big(\trajfeat_D-\trajfeat(\xtraj_R)\big)
 \enspace.
\end{gather}
}
We note that due to the coupling in $\hat\weight'$, the solution to \eqref{eq:theta_update} is non-analytic and can instead be obtained via numerical approaches like Newton-Raphson or quasi-Newton methods.

In previous objective-learning approaches including \cite{bajcsy2017phri} and \cite{maxent}, it is implicitly assumed that all human actions are fully explainable by the robot's representation of the objective function space ($E=1$), leading to the simplified update
\begin{equation}\label{eq:corl_update}
\hat\weight' = \hat\weight - \alpha
\big(\trajfeat_{\change{D}}-\trajfeat(\xtraj_R)\big)
\enspace,
\end{equation}
which can be easily seen to be a special case of \eqref{eq:theta_update} when $P(E=0 \mid \hat\beta)\equiv 0$.
Our proposed update rule therefore \emph{generalizes} commonly-used objective-learning formulations to cases where the human's underlying objective function is not fully captured by the robot's model.
We expect that this extended formulation will enable learning that is more robust to misspecified or incomplete human objective parameterizations.\change{\footnote{\change{Note that to enforce the constraint on $||\theta||=1$, we can indeed project the resulting $\hat{\theta}'$ onto the unit ball. In practice, because our learning from corrections algorithm separates the $\beta$-inference from the $\theta$-inference, this projection is no longer required, but we found it helpful to still constrain the space of $\Theta$ to encourage smoothness in the change of the cost function.}}} Once we obtain the $\hat{\weight}'$ update, we replan the robot trajectory in its 7-DOF configuration space with \change{an} off-the-shelf trajectory optimizer, TrajOpt \cite{trajopt}. 

The update rule changes the weights in the objective in the direction of the feature difference as well, but how much it does so depends on the probability assigned to the correction being well-explained. \change{Looking back at Section \ref{sec:problem}, this update is approximating \eqref{eq:optimize_without_confidence}.} At one extreme, if we know with full certainty that the correction is well explained, then we do the full update as in traditional objective learning. But crucially, at the other extreme, if we know that the correction is poorly explained, we do not update at all and keep our prior belief. 

\begin{algorithm}
\change{
    \caption{\change{Learning from Corrections (Online)}}
    \label{alg:corrections}
    \begin{algorithmic}
        \REQUIRE $P(\hat\beta \mid E=i), \forall i\in\{0,1\}$ from training data.
        \STATE Initialize $\mathbf{x}_R \gets TrajOpt(\hat\weight)$ for initial $\hat\weight$.
        \WHILE{goal not reached}
            \IF{$\control_H \neq \textbf{0}$}
            \STATE $\xtraj_D = \xtraj_R + \mu A^{-1}\tilde\utraj_H$\enspace.
            \STATE $\control^*_H \gets OptimalHumanAction(\Phi_D)$, as per \eqref{opt:optimal_uH}\enspace.
            \STATE $\hat{\beta} = \frac{k}{2\lambda(\|\control_H\|^2-\|\control_H^*\|^2)}\enspace.$
            \STATE $\hat{\theta} \gets \hat\weight - \alpha \frac{\Gamma(\trajfeat_D, E=1)}
 {\Gamma(\trajfeat_D, E=1) +
 \Gamma(\trajfeat_D, E=0)}
 \big(\trajfeat_D-\trajfeat(\xtraj_R)\big)$.
            \STATE $\mathbf{x}_R \gets TrajOpt(\hat\weight)\enspace.$
            \ENDIF
        \ENDWHILE
    \end{algorithmic}
}
\end{algorithm}

\change{Overall, the full algorithm is given in Algorithm  \ref{alg:corrections}. The robot begins tracking a trajectory $\xtraj_R$ given by an initial $\hat\weight$. Once a human torque $\control_H$ is sensed, the robot deforms its trajectory to compute the induced features $\Phi_D$, computes the optimal human action $\control_H^*$ using \eqref{opt:optimal_uH}, and uses it to estimate $\hat\beta$ for that input. It then updates $\hat\weight$ using the learned distributions $P(\hat\beta \mid E=i), \forall i\in\{0,1\}$, and updates its tracked trajectory $\xtraj_R$. For more practical details on how replanning works, and how to set various hyperparameters, consult Appendix \ref{app:practical_corrections}.}

\subsection{Examples}

\change{As in Section \ref{sec:demonstrations}, we now illustrate some examples to help lay out some of the setup we will present in our actual experiments in Sections \ref{sec:case_study} and \ref{sec:study}.} We provide intuition for how the estimators of $\beta$ and $\weight$ work when we have a perfectly specified objective space and a misspecified objective space. In all of the examples, the robot reasons about the previously described distance from the table feature. What changes is the feature for which the human decides to provide corrections.

We look at two situations: the human may correct the relevant feature and push the robot closer to the table, or she might provide an poorly-explained input to keep the coffee mug upright. Fig. \ref{fig:phri_examples} illustrates the two scenarios and contrasts our estimated-$\beta$ approach to the state of the art fixed-$\beta$ approach that uses \eqref{eq:corl_update}.

On the top we present the fixed-$\beta$ method and its performance with both the well-explained and the poorly-explained input. When the input is well explained, the left image shows that the robot learns from the interactions and converges close to the true $\weight=1$. However, when the input is poorly explained on the right, the robot incorrectly learns fictitious $\weight$ values and produces oscillatory behavior.

In the bottom row of Fig. \ref{fig:phri_examples} we present our described estimated-$\beta$ method. In the case of well-explained inputs, the value for $\hat\beta$ increases, allowing $\weight$ to grow up to the real value $\weight=1$. The method has the same behavior as the state of the art. However, more importantly, in the case of poorly-explained input, our method immediately estimates low $\hat\beta$ values, which allows it to significantly reduce unintended learning as compared to the state of the art.

This figure illustrates how situational confidence estimation can aid the robot when the human input is poorly explained. We stress that although our method does not allow the robot to magically learn the correct behavior that the user desires, it greatly reduces unintended learning and undesired behaviors.
\section{Case Studies}
\label{sec:case_study}
Equipped with our algorithmic approaches to situational confidence estimation, we now consider two case studies in learning from demonstrations and corrections using real human input on a 7-DoF robot manipulator.


\subsection{Demonstrations}
We collected human demonstrations of household \change{motion planning} tasks and performed our situational confidence inference offline. 
We recruited 12 people to physically interact with a JACO 7-DoF robotic arm and analyzed 4 common cases that can arise in the context of personal robotics learning.
    
For all the experiments in this section, we asked the participants to provide demonstrations with respect to a feature of interest, which the robot might (well-explained) or might not (poorly-explained) have in its hypothesis space. Some of the features that the humans had to prioritize include: distance of the end effector from the table, distance from the person, or distance from the end-effector to a laptop placed on the table. 
    
Before giving any demonstrations, each person was allowed a period of training with the robot in gravity compensation mode to get accustomed to interacting with the robot. When collecting human demonstrations, participants were asked to move the robot arm holding a cup of coffee from the upper shelf of a cupboard to right above the table, across a laptop.

After collecting all demonstrations, we designed the robot's hypothesis space for inference purposes. In all four scenarios that we will illustrate, the robot reasons over \change{the same three features as in \eqref{eq:features}: E, T, and L.}
\strike{three features: efficiency in the form of sum of squared velocities across the trajectory, distance from the table, and distance from the laptop.}
Although the robot always knows about these features, the demonstrations may have been given relative to different (and potentially unmodeled) features.

Throughout our scenarios, we tested two hypotheses:

\textbf{H1.} \textit{If the human input is well-explained, our inference procedure places high probability on the correct $\weight$ hypothesis, with a high situational confidence $\beta$.}

\textbf{H2.} \textit{If the human input is poorly-explained, our inference procedure does not place high probability on any $\weight$ hypothesis and is uniform over all hypotheses with low situational confidence $\beta$.} 
    
To test these hypotheses, we looked at the resulting inferred belief. Given the demonstrations and a parametrization of the cost function, we first updated the belief over the weight and situational confidence parameters for each single demonstration, $b_{single}(\weight,\beta)$. This gives insights into how a single demonstration can affect the robot's inference procedure. 

Next, we used all 12 human demonstrations to obtain a probability distribution over the weight and confidence measures, $b_{all}(\weight, \beta)$ for each scenario. By using multiple demonstrations as evidence about the cost and the situational confidence parameter, we see how in some scenarios multiple demonstrations can help improve confidence in the $\weight$ estimation. 

We now present experimental results in two scenarios that support our above hypotheses.
    
\subsubsection{Well-specified objective space}
\label{sec:wellspecified}
\begin{figure}[t!]
    \centering
    \includegraphics[width=\columnwidth]{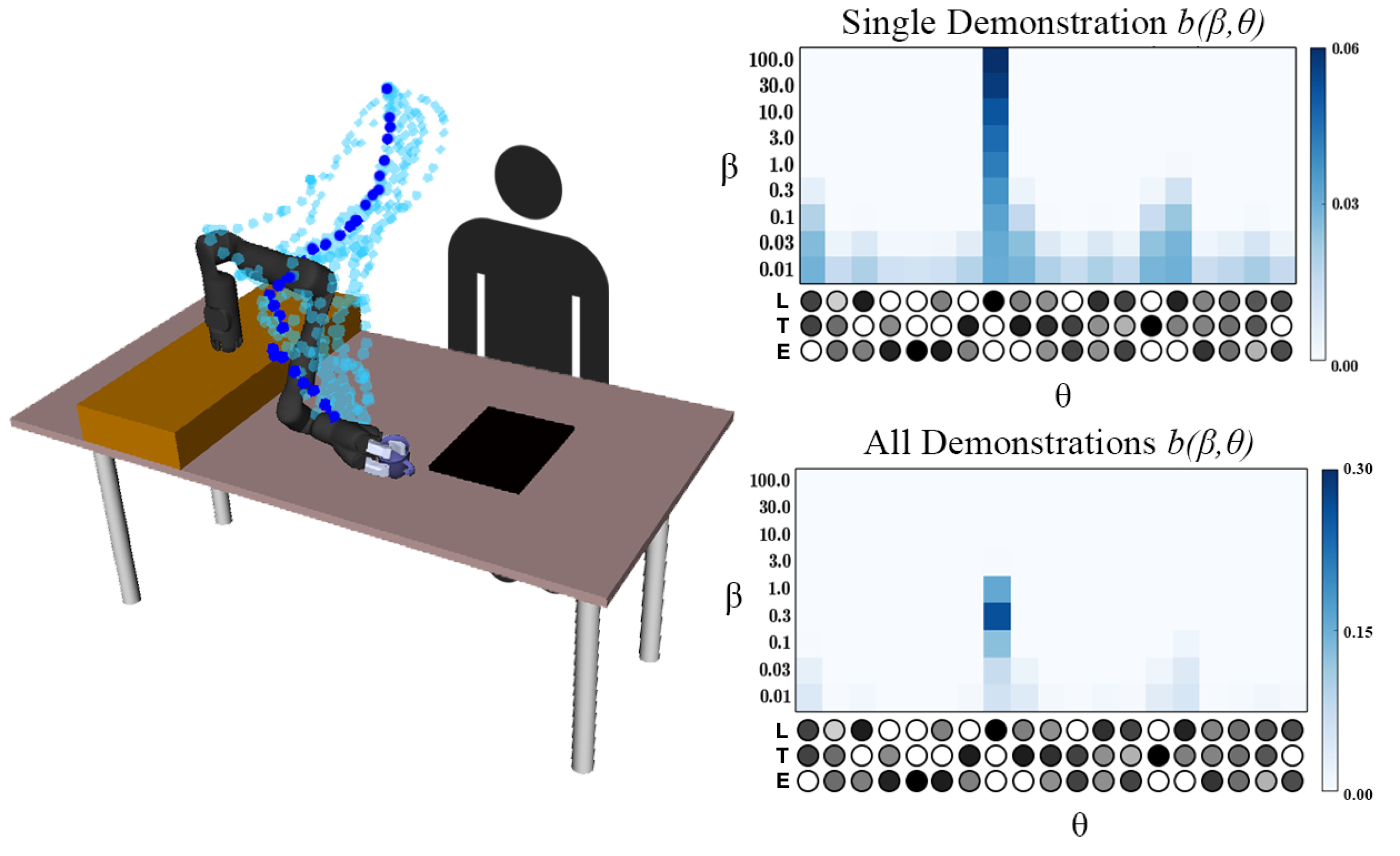}
    \caption{(Left) Human demonstrations avoiding the laptop. (Right) Upper distribution is the posterior belief for the highlighted blue demonstration. Since the robot has the laptop feature in its hypothesis space, this demonstration induces a high $\beta$ on the correct $\weight=[0,0,1]$. Below, when considering all the demonstrations, the inference procedure converges to a slightly lower $\beta$ value due to the suboptimality of some of the demonstrations in the dataset.}
    \label{fig:case1_easy}
\end{figure}

Here we consider a scenario where the robot and the human share the same hypothesis space, i.e. the robot's model is well specified. The participants were instructed to avoid spilling the coffee over the laptop by providing a demonstration where the robot's end-effector is away from the electronic device. Here, the feature of interest was the distance from the laptop which was in the robot's hypothesis space: the demonstration would be well explained as long as the demonstration maintained a distance of at least $L$ meters away from the center of the laptop.

On the left of Fig. \ref{fig:case1_easy} we  visualize all 12 recorded demonstrations and the experimental setup. Note that most participants had an easy time providing demonstrations which avoided the laptop. Indeed, we noticed that 10 out of the 12 demonstrations resulted in high situational confidence and a probability distribution similar to the one at the top right of Fig. \ref{fig:case1_easy}. Here, the $\weight$ vector that has largest weight on the third feature (distance from the laptop) is correctly inferred to have high $\beta$ value. This signals that the robot is highly confident the person provided a demonstration that avoids the laptop, which is correct and supports our hypothesis H1.

Another interesting observation is that the situational confidence over all 12 demonstrations together is lower than in the case of the single optimal demonstration highlighted in blue (peak at around 1.0 instead of 100.0)\footnote{In the lower right belief in Fig. \ref{fig:case1_easy}, note from the colorbar values that the probability mass is more peaked than in the case of a single demonstration. This confirms our intuition that the robot's certainty in the hypothesis is enhanced the more demonstrations supporting that hypothesis it receives.}. This is due to the two noisy demonstrations that came too close to the laptop. When working with non-expert users, it is inevitable that such imperfect demonstrations will arise. However, despite the challenge of noisy and/or erroneous demonstrations, the algorithm recovers the correct $\weight$ hypothesis with a relatively high $\beta$, supporting H1 once again.

\subsubsection{Misspecified hypothesis space}
\label{sec:misspecification}
\begin{figure}[t!]
    \centering
    \includegraphics[width=\columnwidth]{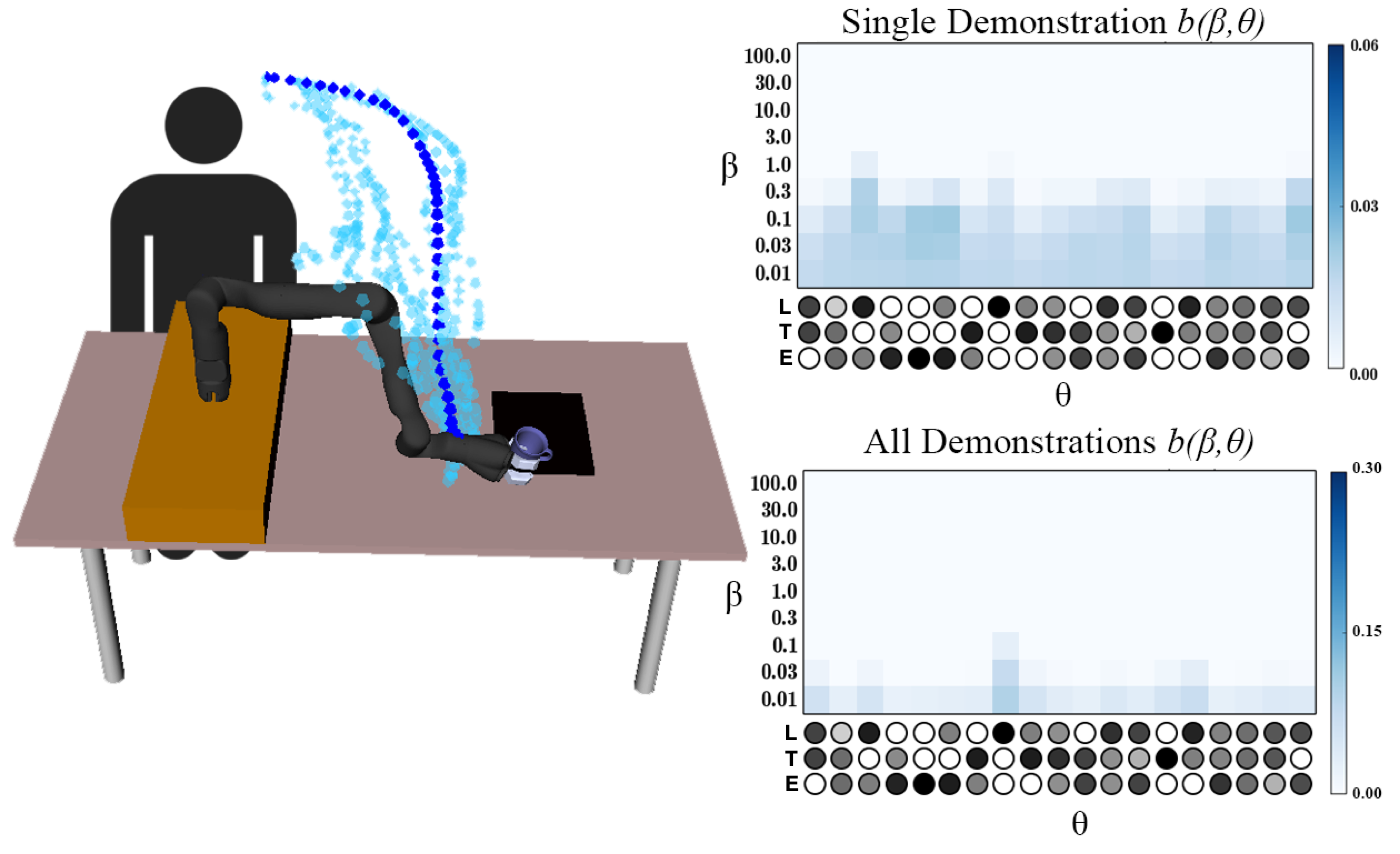}
    \caption{(Left) Human demonstrations avoiding the user's body. (Right) Upper distribution is the posterior belief $b(\beta,\weight)$ for the highlighted demonstration. Since the robot's model does not include distance to the user's body, none of the robot's hypotheses can explain the demonstration, as reflected in the higher probabilities on low $\beta$s for all $\weight$s. After performing inference on all the demonstrations, the distribution in the lower right plot shows even more probability mass on the lowest situational confidence values.}
    \label{fig:case1_misspecified}
\end{figure}

We look at the opposite scenario: the robot and the human do not share the same hypothesis space and the robot's model is clearly misspecified. 

Participants were instructed to move the robot from start to end while also keeping the robot's hand away from their body to avoid spilling coffee on their clothes. Since the robot's cost function does not include any notion of distance to humans, the demonstrations should appear poorly explained relative to the robot's model of how humans choose demonstrations. 

Fig. \ref{fig:case1_misspecified} visualizes all 12 demonstrations as well as the posterior probability distributions for a single highlighted trajectory and for all 12. For both a single demonstration and all of them, in the case of misspecification none of the hypotheses are correct. Thus, the robot infers equally low probability for all $\weight$s, with low situational confidence, supporting our hypothesis H2. This signals that the robot is unsure what the person's demonstration referred to, as we expected.
    
These two examples illustrate cases where our method supports the two hypotheses above. However, there are important limitations that we discuss in the following two scenarios. 
    
\subsubsection{Feature correlation}
\label{sec:feat_correlation}
\begin{figure}[t!]
    \centering
    \includegraphics[width=\columnwidth]{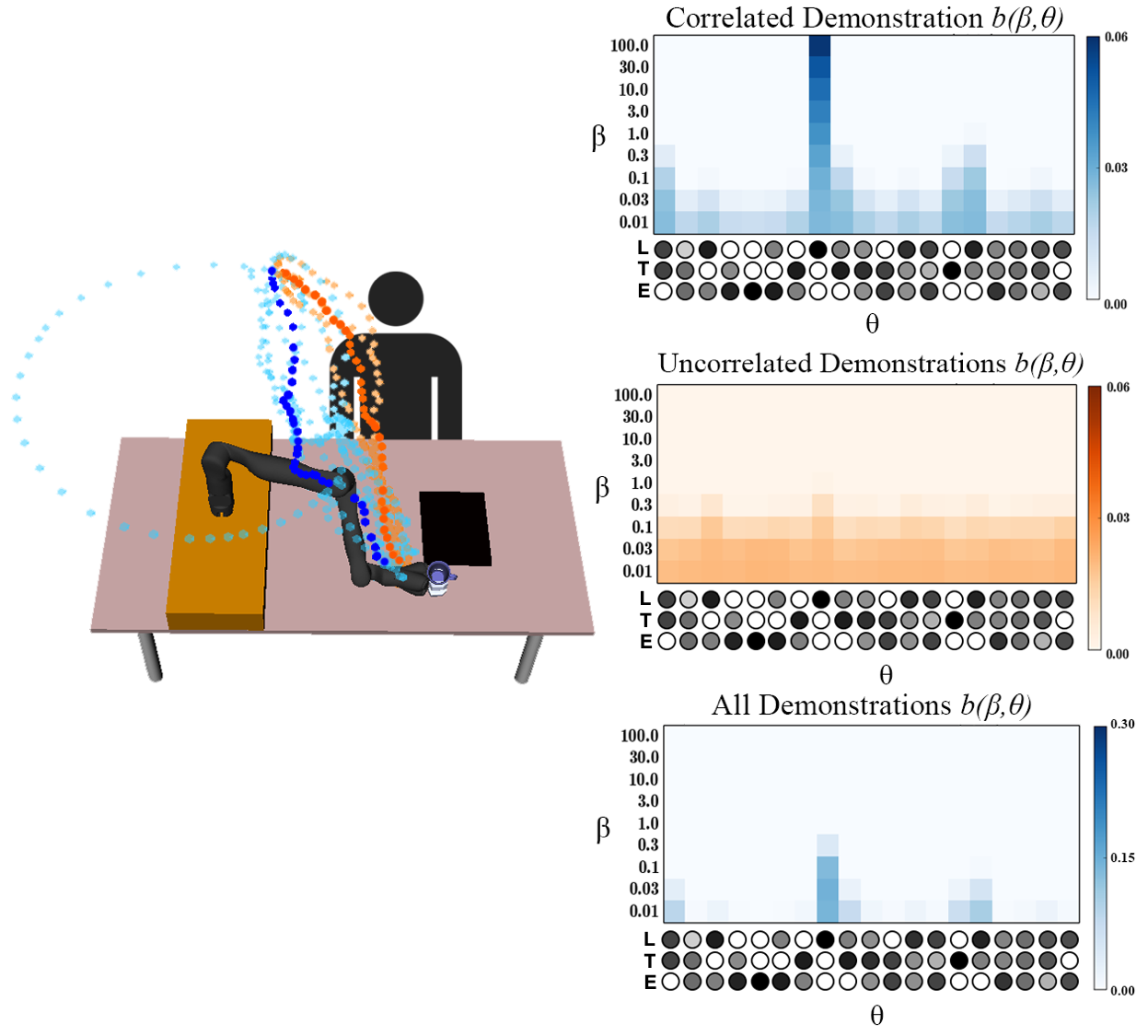}
    \caption{(Left) Human demonstrations avoiding the user's body. The blue cluster is correlated with the feature describing distance from the laptop. The orange cluster is uncorrelated. (Right) The top distribution is the posterior belief $b(\beta,\theta)$ for the highlighted blue correlated demonstration. Notice that the hypothesis that puts all weight on avoiding the laptop $\weight=[0,0,1]$ dominates the distribution. Meanwhile, the posterior belief for the highlighted orange demonstration indicates low situational confidence in all hypotheses. The bottom distribution shows that when combining all demonstrations, the robot continues to have low situational confidence although the laptop hypothesis has slightly higher $\beta$.}
    \label{fig:case2_correlation}
\end{figure}

The past two examples demonstrate clear instances when the robot's objective space is either well specified or misspecified. However, often times situations will be more ambiguous. For example, although the human input may refer to a feature that is nonexistent in the robot's hypothesis space, the robot may know about a feature that is correlated to the one the human is trying to affect. In this next scenario, we investigate how such feature correlation influences the situational confidence estimates.
        
We asked the participants to move the robot from the same start and end as before, while keeping the cup in the robot's end-effector away from their body to avoid spilling coffee on their clothes. The setup is similar to the poorly-explained demonstration in the previous scenario, only that now the human starts in a different initial position.
        
Visualizations of the 12 demonstrations in Fig. \ref{fig:case2_correlation} showcase that although all demonstrations move the cup away from the person, some of them (depicted in blue) also maintain a good distance away from the laptop. Hence, even though the human was trying to teach the robot to stay away from their body, the robot interprets the human's demonstrations as a signal to stay away from the laptop. Thus, we say that the distance from human and distance from laptop features are \textit{correlated}.
        
When looking at the top-right posterior probability in Fig. \ref{fig:case2_correlation}, the distribution over $\weight,\beta$ shows that our algorithm infers a high situational confidence for the $\weight$ that fully considers the distance from the laptop. Thus, even if the human input does not pertain to a feature in the robot's hypothesis space, in some cases the demonstration can still be explained via correlated features in the robot's hypothesis space. This observation does not support H2 and is clearly a limitation of our method.
        
However, the orange cluster of demonstrations in Fig. \ref{fig:case2_correlation}, showcase the fine line between demonstrations that induce a feature correlation and those that do not. The orange demonstrations clearly ignore the laptop and simply take the shortest path to the end goal while avoiding the human's body. As we can see in the orange probability distribution, our method infers a uniform distribution over all $\weight$ hypotheses, with a focus on the lowest situational confidence values, backing H2.
        
These two clusters highlight that our method infers reasonable $\weight, \beta$ values even in the case of feature correlation. The robot either infers a good $\weight$ to perform its original task through the means of another feature, or it has low confidence in understanding the input.
        
When we look at the posterior distribution that results from all 12 demonstrations, the bottom-right part of the figure shows that, due to the correlation in the blue cluster, there is increased probability on the $\weight$ that considers fully the distance from the laptop. However, due to the ambiguity of the orange cluster, the situational confidence is not as high as it would be in a well-explained case (see Fig. \ref{fig:case1_easy}). 

\subsubsection{Feature engineering}
\label{feature_eng}
\begin{figure}[t!]
    \centering
    \includegraphics[width=\columnwidth]{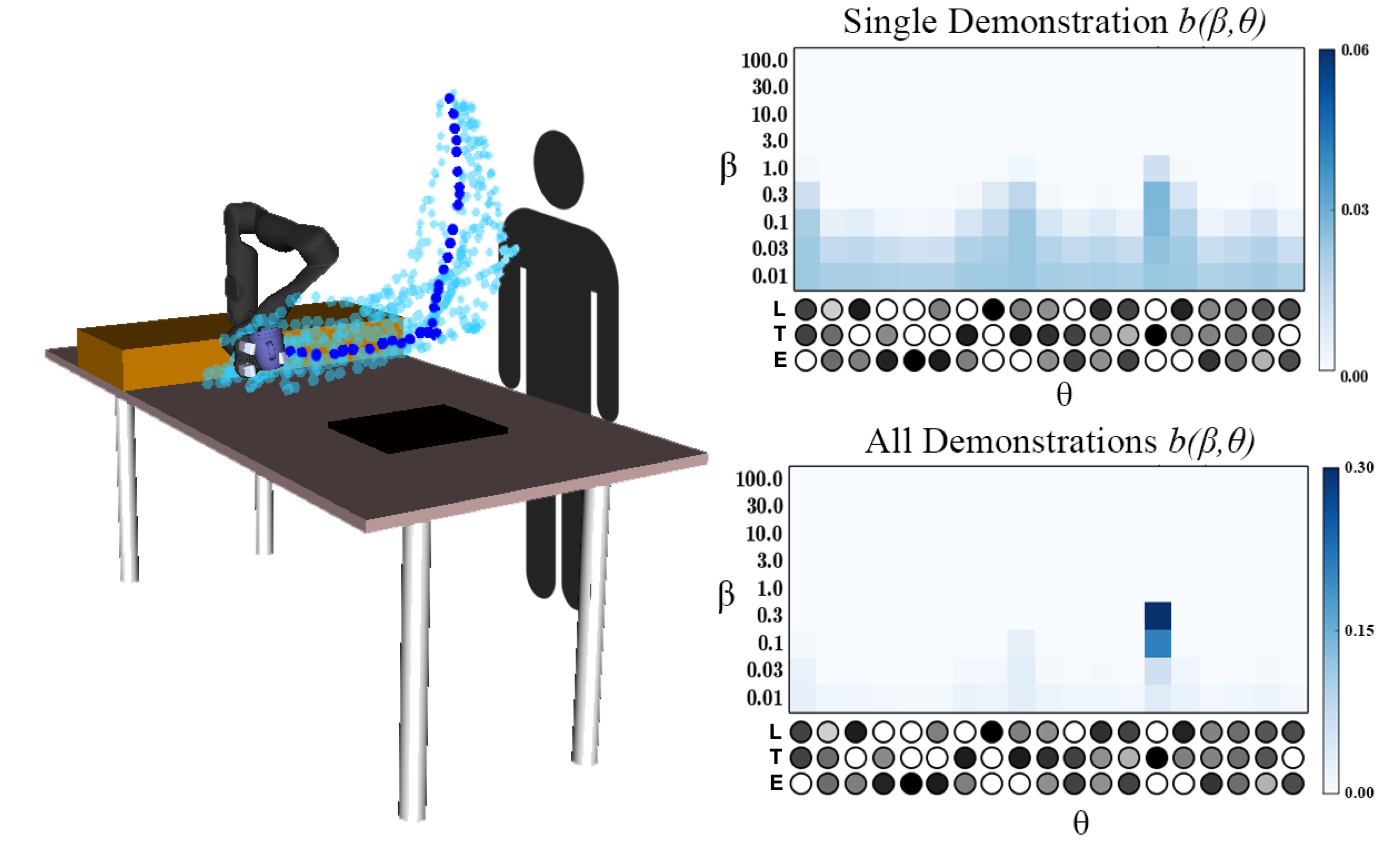}
    \caption{(Left) Human demonstrations keeping the cup in the end-effector close to the table. (Right) Because it is difficult for the person to give a good demonstration, the top posterior does not have a clearly defined peak for one particular hypothesis, although several $\theta$s are favored. In the bottom distribution, we notice that when presented with all 12 demonstrations, the robot can more clearly infer the correct hypothesis for the distance to the table, $\weight=[0,1,0]$.}
    \label{fig:case3_engineering_demo}
\end{figure}

Many of the cost function features we considered so far have been intuitive to provide demonstrations for. However, some cost functions may be particularly challenging or unintuitive for human users. Two extreme examples of this could be features learned using complex function approximators or unintuitive features like minimizing the total energy of a system.

In our scenario, the feature users have a difficult time providing good demonstrations for is the distance between the robot's hand and the table along the trajectory. Since the feature was designed as the sum of distances to table for all waypoints in the trajectory, the optimal demonstration immediately moves the end-effector to the table and then keeps it right above the tabletop for the rest of the path, as seen in Fig. \ref{spec}. This limitation does not support H1.

However, this mathematical optimum does not necessarily align with how human users interpret the best behavior for this task. In our experiments, most users gradually bring the robot's hand closer to the table, rather than pushing it down immediately, for a more smooth and natural motion (see left in Fig. \ref{fig:case3_engineering_demo}). These demonstrations thus appear noisy and sub-optimal with respect to the robot's model and make it difficult to infer the true $\theta$ from a single demonstration. 

This phenomenon is reflected more clearly when we look at the top-right belief distribution in Fig. \ref{fig:case3_engineering_demo}. Although the distribution for the highlighted blue demonstration has some peaks around hypotheses that strongly favor the feature responsible for distance to the table, it is not nearly as clearly defined as it should be for a well-explained demonstration (see Fig. \ref{fig:case1_easy}).

However, when the robot gathers evidence from multiple demonstrations, the algorithm does manage to figure out that this is the feature that people were optimizing for. The bottom right plot in Fig. \ref{fig:case3_engineering_demo} illustrates that, once again, having more input samples eventually leads our algorithm to converge to a strong probability for the right $\weight$ with a reasonably high $\beta$. Although our method cannot back H1 when inferring the objective from a single demonstration, more data leads our algorithm to correctly support H1.

\noindent\textbf{Summary: } The four situations presented above illustrate that our two original hypotheses H1 and H2 are supported most of the time (\ref{sec:wellspecified}, \ref{sec:misspecification}), with some exceptions (\ref{sec:feat_correlation}, \ref{feature_eng}). We saw that when the person has a difficult time giving a good demonstration (Section \ref{feature_eng}), our method cannot support H1 unless provided with multiple demonstrations, to disambiguate the inherent noise in the user's suboptimal input. Additionally, when the person provides what should be a poorly-explained demonstration (Section \ref{sec:feat_correlation}), feature correlation might lead the inference to falsely detect $\weight$s corresponding to that input, contradicting H2. However, we observed that when given more demonstrations, our algorithm can attribute low situational confidence $\beta$ if the uncorrelated input is sufficient. More work is needed in this area.

\subsection{Corrections}
\label{sec:case_study_corrections}

We now turn our attention to case where human input is sparse and in the form of intermediate corrections during the robot's task execution. Here \change{we present an offline case study where} we analyze how our estimates of $\hat{\beta}$ enable us to distinguish if the input is well explained or not to the robot's model of the human. \change{For a full exploration of the real-time updates from human corrections, we conduct an online user study which we later describe in Section \ref{sec:study}.}

We recruited 12 additional individuals to physically interact with the same robotic manipulator. Each participant was asked to intentionally correct a feature (that the robot may or may not have in its hypothesis space): adjusting the distance of the end effector from the table, adjusting the distance from the person, or adjusting the cup's orientation. 

During this case study the robot did not attempt to update the feature weights $\weight$ and simply tracked a predefined trajectory with an impedance controller \cite{impedance}. The participants were instructed to intervene only once during the robot's task execution, in order to record a single physical correction. The resulting trajectories and physical interaction $\control_H$ were saved for offline analysis. This setup enabled us to easily analyze the situational confidence of the robot as we changed the robot's hypothesis space.

Next, we ran our approximate inference algorithm using the recorded human interaction torques and robot joint angle information. We measured what $\hat{\beta}$ would have been for each interaction if the robot knew about a given subset of the features. By changing the subset of features for the robot, we changed whether any given human interaction was well \emph{explained} to the robot's hypothesis space.

We tested two hypotheses:

\textbf{H1.} \textit{Well-explained interactions result in high $\hat{\beta}$, whereas interactions that change a feature the robot \textbf{does not} know about result in low $\hat{\beta}$ for all features the robot \textbf{does} know about.}

\textbf{H2.} \textit{Not reasoning about well-explained interactions and, instead, indiscriminately learning from every update leads to significant unintended learning.}

We ran a repeated-measures ANOVA to test the effect of whether and input is well explained on our $\hat{\beta}$. We found a significant effect ($F(1,521)=9.9093$, $p=0.0017$): when the person was providing a well-explained correction, $\hat{\beta}$ was significantly higher. This supports our hypothesis H1.

\begin{figure}[t!]
\subfloat[Average $\beta$ for well-explained and poorly-explained interactions.]{\label{fig:beta_relevance}\includegraphics[width=\columnwidth]{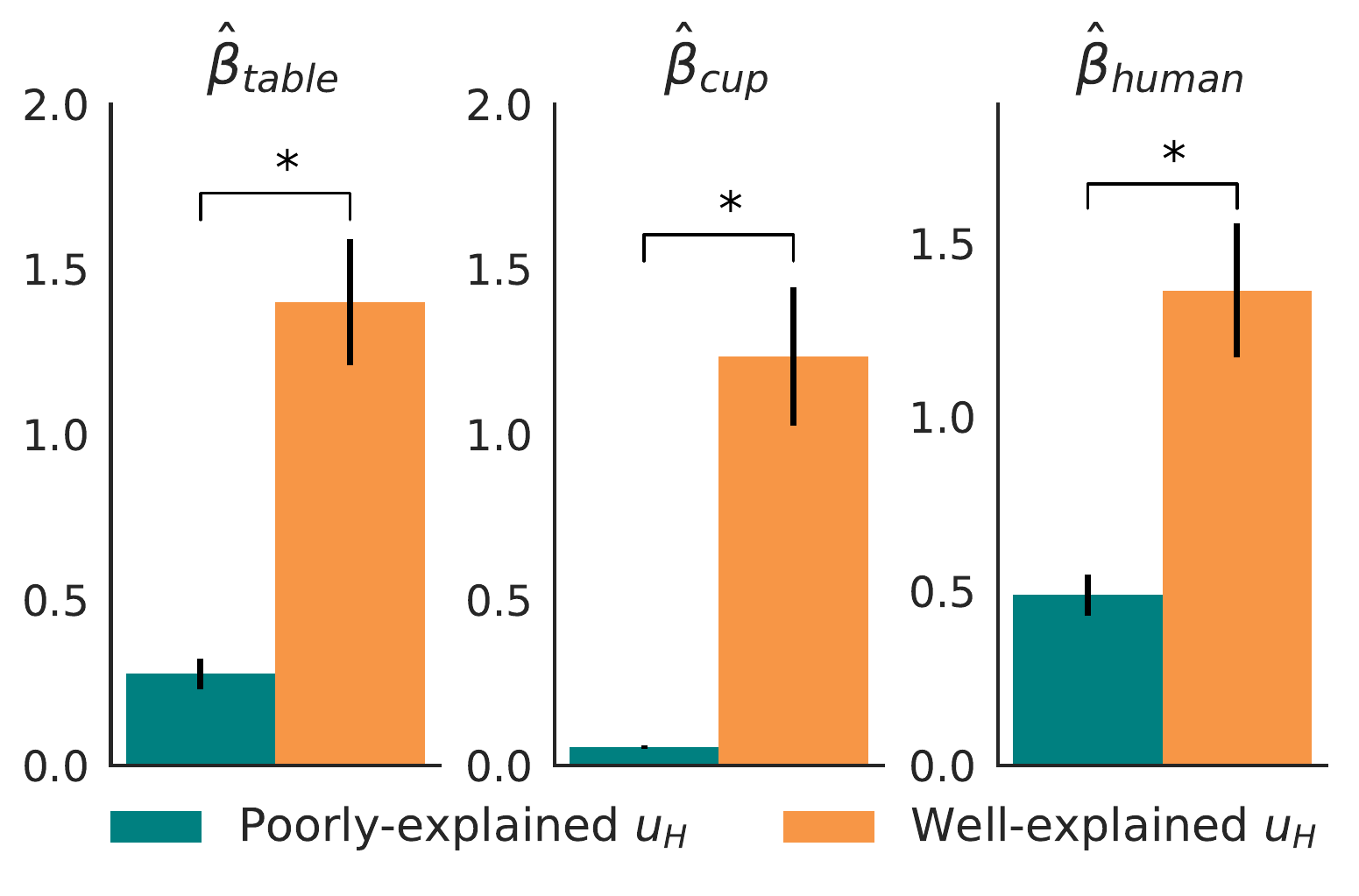}} \\
\subfloat[Average $\Delta\Phi$ for well-explained and poorly-explained interactions.]{\label{fig:update_relevance}\includegraphics[width=\columnwidth]{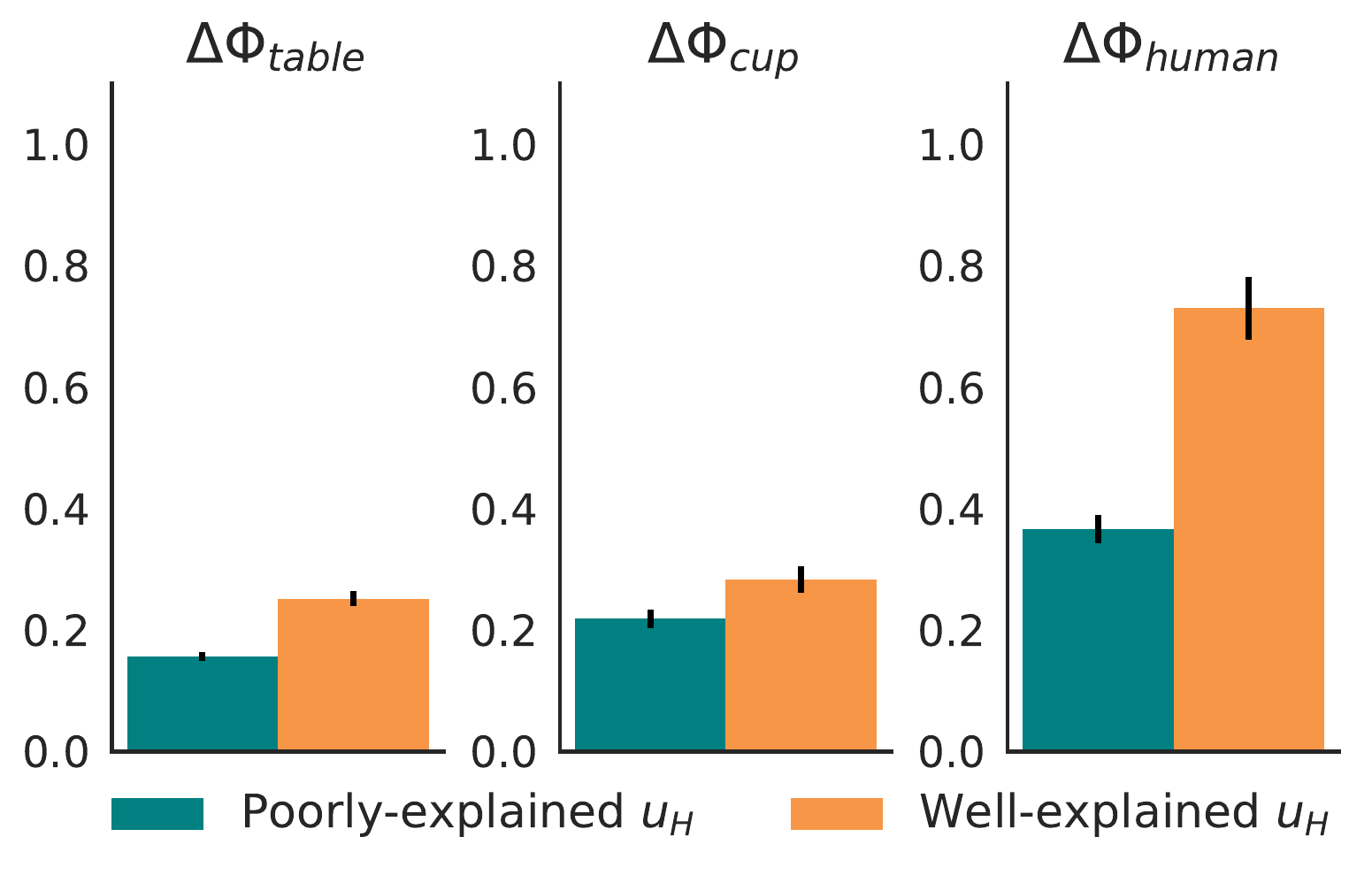}} 
\caption{$\beta$ values are significantly larger for well-explained actions than for poorly-explained ones. Feature updates are non-negligible even during poorly-explained actions, which leads to significant unintended learning for fixed-$\beta$ methods.}
\label{fig:offline_relevant}
\end{figure}

Fig. \ref{fig:beta_relevance} plots $\hat{\beta}$ under the well-explained (orange) and poorly-explained (blue) conditions. Whereas the poorly-explained interactions end up with $\hat{\beta}$s close to 0, well-explained corrections have higher mean and take on a wider range of values, reflecting varying degrees of human performance in correcting something the robot knows about. We fit per-feature chi-squared distributions for $P(\hat{\beta} \mid E)$ for each value of $E$ which we will use to infer $E$ and, thus, $\weight$ online. In addition, Fig. \ref{fig:update_relevance} illustrates that even for poorly-explained human actions $\control_H$, the resulting feature difference $\Delta\Phi = \trajfeat(\xtraj_D)-\trajfeat(\xtraj)$ is non-negligible. This supports our second hypothesis, H2, that not reasoning about how well-explained an action is is detrimental to learning performance when the robot receives misspecified updates.

\section{User Study on Learning from Corrections}
\label{sec:study}

Our case study on corrections suggested that $\hat{\beta}$ can be used as a measure  of whether physical interactions are well explained and should be learned from. Next, we conducted an IRB-approved user study to investigate the implications of using these estimates during learning. During each experimental task, the robot began with a number of incorrect weights and participants were asked to physically correct the robot. Locations of the objects and human were kept consistent in our experiments across tasks and users to control for confounds\footnote{We assume full observability of where the objects and the human are, as the focus of this paper is not sensing.}. The planning and inference were done for robot trajectories in 7-dimensional configuration space, accounting for all relevant constraints including joint limits and self-collisions, as well as collisions between obstacles in the workspace and any part of the robot’s body.\footnote{For video footage of the experiment, see: \href{https://youtu.be/stnFye8HdcU}{https://youtu.be/stnFye8HdcU}} 

\subsection{Experiment design}

\subsubsection{Independent variables}
We used a 2 by 2 factoral design. We manipulated the corrections learning strategy with two levels (fixed-$\beta$ and estimated-$\beta$ learning), and also whether the human corrected for features inside (well explained) or outside (poorly explained) the robot's hypothesis space. In the fixed learning strategy, the robot updated its feature weights from a given interaction via \eqref{eq:corl_update} with a fixed $\beta$ value. In the estimated-$\beta$ learning strategy, the robot updates its feature weights via \eqref{eq:theta_update}. The offline experiments above provided us an estimate for $P(E\mid \hat{\beta})$ that we used in the gradient update. 

\subsubsection{Dependent measures - objective}

To analyze the objective performance of the two learning strategies, we focused on comparing two main measurements: the length of the $\hat{\weight}$ path through weight space as a measurement of the learning process, and the regret in feature space measured by
$|\trajfeat(\xtraj_{\weight^*}) - \trajfeat(\xtraj_{actual})|$. \change{Longer $\hat{\weight}$ paths should indicate a learning process that oscillates, whereas shorter paths suggest smoother learning curves. On the other hand, high regret implies that the learning method did not converge to a good objective $\theta$, whereas low regret indicates better learning.}

\subsubsection{Dependent measures - subjective} For each condition, we administered a 7-point Likert scale survey about the participant's interaction experience (Table \ref{tab:likert}). We separate the survey into 3 scales: task completion, task understanding, and unintended learning.


\subsubsection{Hypotheses} We tested four hypotheses:

\textbf{H1.} \textit{On tasks where humans try to correct \textbf{inside} the robot's hypothesis space (well-explained corrections), inferring situational confidence is not inferior to always assuming high situational confidence.}

\textbf{H2.} \textit{On tasks where humans try to correct \textbf{outside} the robot's hypothesis space (poorly-explained corrections), inferring situational confidence reduces unintended learning.}

\textbf{H3.} \textit{On tasks where they tried to correct \textbf{inside} the robot's hypothesis space, participants felt like the two methods performed the same.}

\textbf{H4.} \textit{On tasks where they tried to correct \textbf{outside} the robot's hypothesis space, participants felt like our estimated-$\beta$ method reduced unintended learning.}

\subsubsection{Tasks}
We designed 4 experimental household \strike{manipulation} \change{motion planning} tasks for the robot to perform in a shared workspace. Similarly to the case studies, for each experimental task, the robot carried a cup from a start to end pose with an initially incorrect objective. Participants were instructed to physically intervene to correct the robot's behavior during the task. 

In Tasks 1 and 2, the robot's default trajectory took a cup from the participant and put it down on the table, but carried the cup too high above the table. In Tasks 3 and 4, the robot also took a cup from the human and placed it on the table, but this time it initially grasped the cup at the wrong angle, requiring human assistance to correct end-effector orientation to an upright position. For Tasks 1 and 3, the robot knew about the feature the human was asked to correct for ($E=1$) and participants were told that the robot should be compliant. For Tasks 2 and 4, the correction was poorly explained ($E=0$) and participants were instructed to correct any additional unwanted changes in the trajectory.

\subsubsection{Participants}
We used a within-subjects design and randomized the order of the learning methods during experiments. We recruited 12 participants (6 females, 6 males, aged 18-30) from the campus community, 10 of which had technical background. None of the participants had experience interacting with the robot used in our experiments.

\begin{figure}[t!]
\subfloat[Regret averaged across subjects.]{\label{fig:regret}\includegraphics[width=\columnwidth]{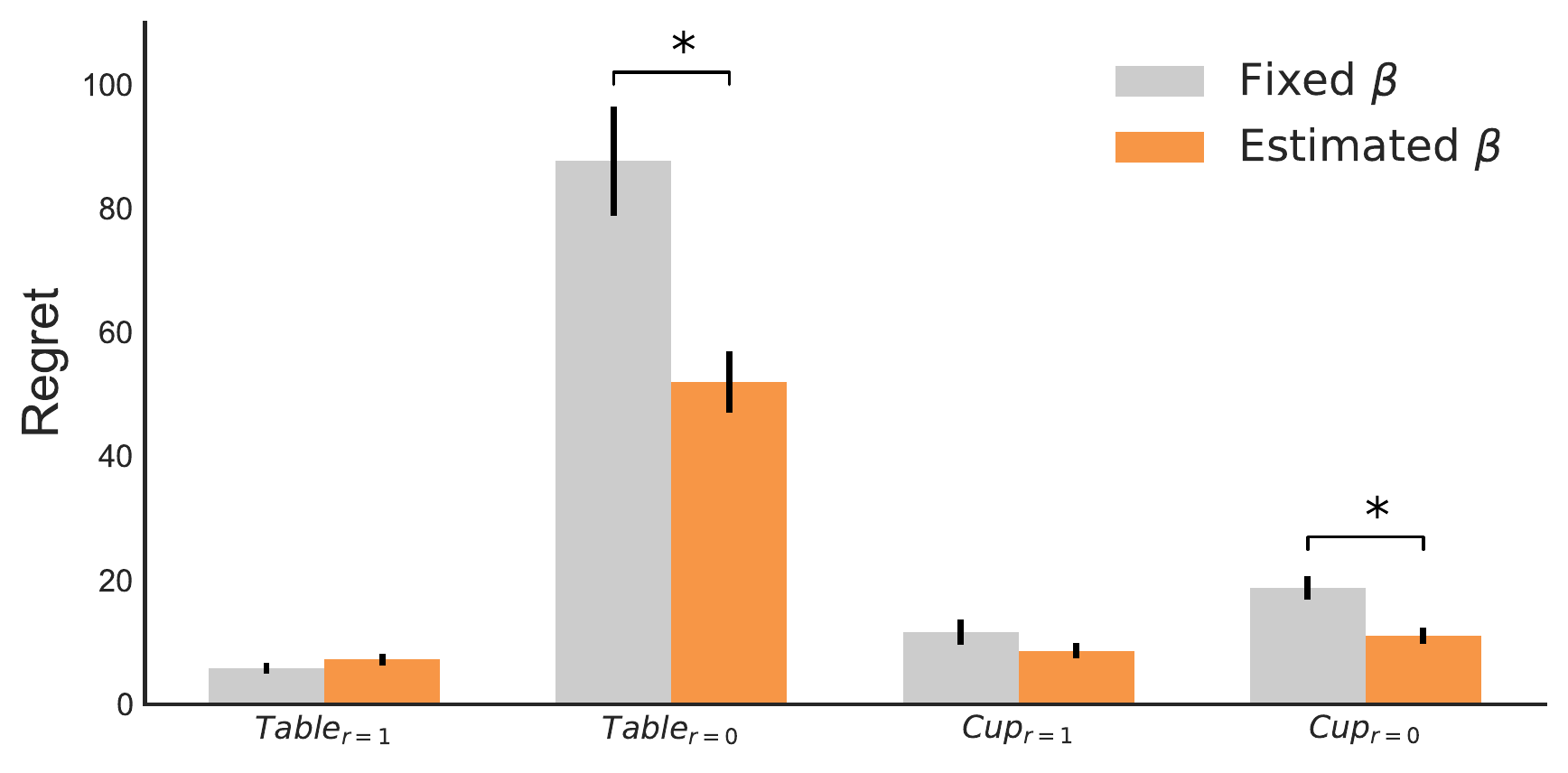}} \\
\subfloat[$\hat{\theta}$ learning path length averaged across subjects.]{\label{fig:weight_path}\includegraphics[width=\columnwidth]{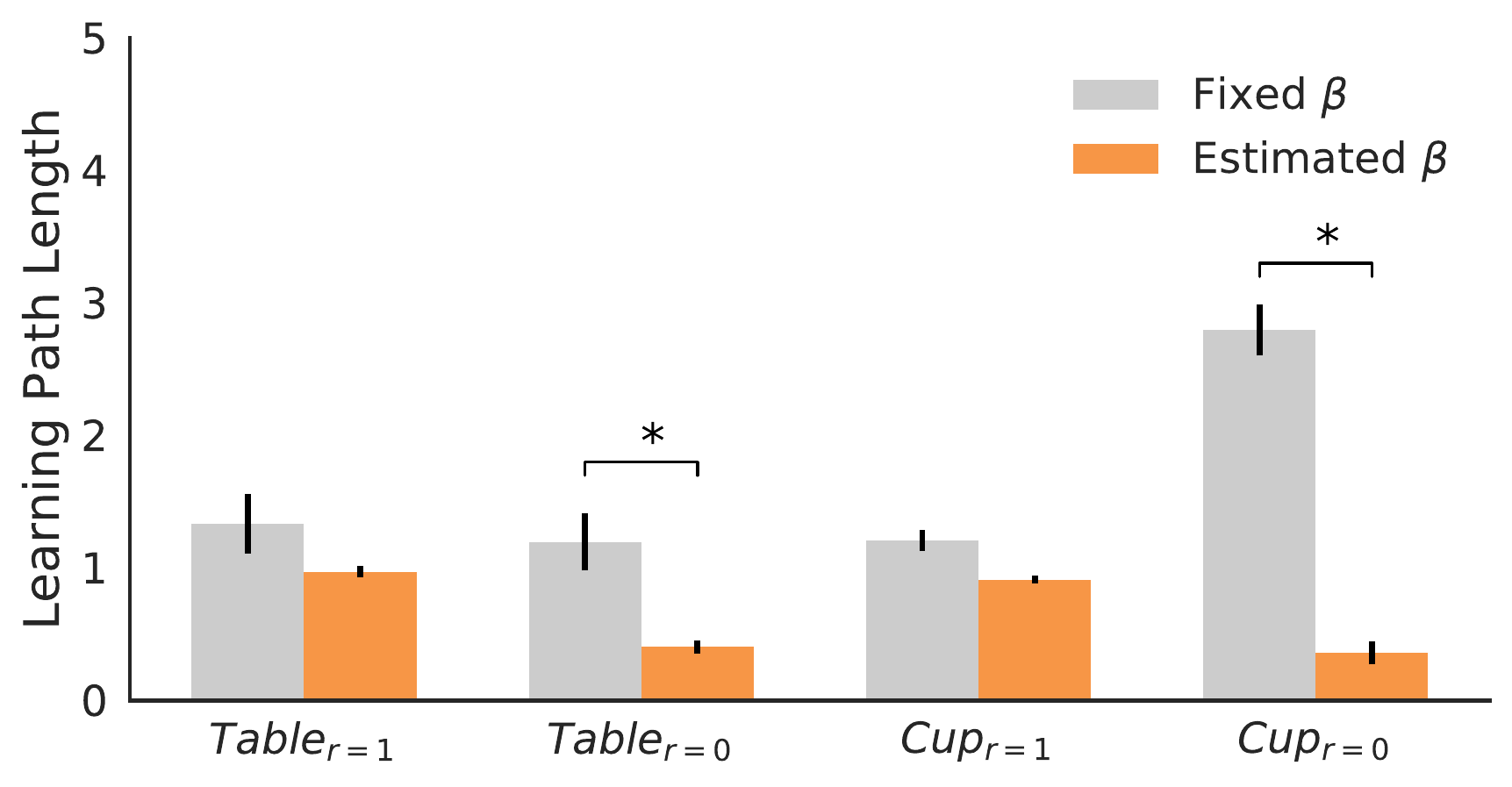}} \caption{Comparison of regret and length of $\hat{\theta}$ learning path through weight space over time (lower is better).
}
\label{fig:delta_cost}
\end{figure}

{\renewcommand{\arraystretch}{1.5}
\begin{table*}[t!]
 \centering
 \adjustbox{max width=\textwidth}{
 \begin{tabular}{|c|l|c|c|c|}
 \hline
 & \multicolumn{1}{c|}{\textbf{Questions}} & \multicolumn{1}{c|}{\textbf{Cronbach's $\alpha$}} & \multicolumn{1}{c|}{\textbf{F-Ratio}} & \multicolumn{1}{c|}{\textbf{p-value}} \\
 \hline
 \parbox[t]{2mm}{\multirow{2}{*}{\rotatebox[origin=c]{90}{\textbf{task}}}} 
 & The robot accomplished the task in the way I wanted. & \multirow{2}{*}{0.94} & \multirow{2}{*}{0.88} & \multirow{2}{*}{0.348}\\
 & The robot was NOT able to complete the task correctly.  &&&\\
 \hline
 \parbox[t]{2mm}{\multirow{4}{*}{\rotatebox[origin=c]{90}{\textbf{understand}}}} 
 &&&&\\
 & I felt the robot understood how I wanted the task done. & \multirow{2}{*}{0.95} & \multirow{2}{*}{0.55} & \multirow{2}{*}{0.46}\\
 &  I felt the robot did NOT know how I wanted the task done. &&&\\
 &&&&\\
 \hline
 \parbox[t]{2mm}{\multirow{4}{*}{\rotatebox[origin=c]{90}{\textbf{unintend}}}} 
 & I had to undo corrections that I gave the robot. &&&\\
 & The robot wrongly updated its understanding about aspects of the task I did not want to change.  & \multirow{2}{*}{0.91} & \multirow{2}{*}{9.15} & \multirow{2}{*}{\textbf{0.0046}}\\
 & After I adjusted the robot, it continued to do the other parts of the task correctly.  &&&\\
 & After I adjusted the robot, it incorrectly updated parts of the task that were already correct. &&&\\
 \hline
 \end{tabular}
 }
 \strut 
 \label{table}
 \caption{Results of ANOVA on subjective metrics collected from a 7-point Likert-scale survey.} \label{tab:likert}
 \end{table*}}

\subsubsection{Procedure}
Every participant was assigned a random ordering of the two methods, and performed each task without knowing how the underlying methods work. \change{One challenge in performing and evaluating our experiment was that different participants may have different internal preferences for how a task should be performed. In order to have a consistent notion of ground-truth preferences, we fixed the true objective (e.g. how far the cup should be from the table) for each task.} At the beginning of each task, the participant was first shown the incorrect default trajectory that they must correct, followed by the ground-truth desired trajectory they should teach the robot. \change{This allows us to focus only on how well each algorithm infers objectives from human input, versus trying to additionally estimate the unique ground-truth human objective of each participant.} Then the participant performed a familiarization round, followed by two recorded experimental rounds. After answering the survey, the participant repeated the procedure for the other method. 

\subsection{Analysis}

\subsubsection{Objective}
We ran a repeated-measures factorial ANOVA with learning strategy and input quality (well or poorly explained) as factors for the regret. We found a significant main effect for the method ($F(1,187)=7.8, p=0.0058$), and a significant interaction effect ($F(1,187)=6.77, p=0.0101$). We ran a post-hoc analysis with Tukey HSD corrections for multiple comparisons to analyze this effect, and found that it supported our hypotheses. On tasks where corrections were poorly explained, the estimated-$\beta$ method had significantly lower regret ($p=0.001$); on tasks where corrections were well explained, there was no significant difference ($p=0.9991$). Fig. \ref{fig:regret} plots the regret per task, and indeed the estimated-$\beta$ method was not inferior on tasks 1 and 3, and significantly better on tasks 2 and 4. 

For the length of the $\hat\weight$ path through weight space metric, the factorial ANOVA analysis found a significant main effect for the method ($F(1,187)=76.43, p<0.0001$), and a significant interaction effect ($F(1,187)=33.3, p<0.0001$). A similar post-hoc analysis with Tukey HSD correction for multiple comparisons also supports our hypotheses. On tasks where corrections were poorly explained, our method had significantly lower average weight paths over time ($p=0.0025$); on tasks where correction were well explained, however, there was no significant difference ($p=0.1584$). The same results are supported by Fig. \ref{fig:weight_path}, which plots the average length of $\hat{\theta}$ through weight space per task, and indeed our method was not significantly inferior for tasks 1 and 3, and significantly better on tasks 2 and 4. 

\subsubsection{Subjective}
We ran a repeated measures ANOVA on the results of our participant survey. We find that our method is not significantly different from the baseline in terms of task completion ($F(1,7)=0.88,p=0.348$) and task understanding ($F(1,7)=0.55, p=0.46$), which supports H3. Participants also significantly preferred the estimated-$\beta$ method in terms of reducing unintended learning ($F(1,7)=9.15,p=0.0046$), which supports H4.

\section{Discussion}
\label{sec:conclusion}

Human guidance is becoming increasingly important as autonomous systems enter the real world. One common way for robots to interpret human input is treating it as evidence about hypotheses in the robot's objective space. Since accounting for all possible hypotheses and situations ahead of time is challenging if not infeasible, in this paper we claim that robots should explicitly reason about how well their given hypothesis space can explain the human inputs. 

We introduced the notion of \textit{situational confidence} $\beta$ as a natural way to measure how much the robot should trust its inputs and \strike{in turn }learn from them. We presented a general \change{framework} for estimating $\beta$ in conjunction with any task objectives for scenarios where the human and the robot are operating the same dynamical system. \change{We instantiated it} for learning from human demonstrations, as well as \change{for learning from corrections, by deriving a close to real-time} approximate algorithm. In both settings, we exemplified -- via human experiments with a 7-DoF robotic manipulator and a user study -- that reasoning about situational confidence does, in fact, assist the robot in better understanding when it cannot explain human input.

There are several important limitations in our work.
\change{Perhaps the biggest limitation of all, which we alluded to in Section \ref{sec:intro}, is that the hypothesis space can be misspecified but the robot can nonetheless explain the input relatively well, thus confusing misspecification for slight noise. This is especially true in more expressive hypothesis spaces, where there might always be some hypothesis that explains the input. This is unfortunately a fundamental problem with detecting misspecification in expressive hypothesis spaces: a single demonstration or a single data point will not be enough. Much like learning cost functions when using such spaces requires much more and diverse data than when using a less expressive space, with detecting misspecification too it will be the case that the robot will require a rich and diverse set of data points. The more data the robot has access to, and the more diversely it is distributed, the less of a chance there is that one wrong hypothesis can explain all the data.}

\change{Furthermore, our approach cannot disambiguate between misspecification of the hypothesis space and misspecification of the human observation model, i.e. the Boltzmann model.}

\strike{In order to interpret human input, throughout the paper we have modeled the human's choice of control as an independent, random draw from a Boltzmann distribution. While this distributional assumption with roots in econometrics and cognitive science is becoming more popular in robotics, it is by no means a universally accurate model of human decision making.} 

\strike{Additionally, we used hand-designed features throughout the experiments. In Section \mbox{\ref{feature_eng}} we saw that these features can be challenging to design. Instead, the robot could use data-driven techniques, such as function approximators, to obtain richer models of the cost function. However, learning from a single human input in general becomes problematic due to the very limited training set compared to the space of possible feature misspecifications.}

\change{Algorithmically, while for corrections we derived a way to handle continuous hypothesis spaces that scales linearly with the dimensionality of the space, for demonstrations we relied on simply discretizing the space. This was sufficient for showcasing the benefit of estimating situational confidence, since for demonstrations this is done offline. However, to scale the method to complex spaces, we need to combine it with state-of-the-art (Bayesian) IRL approaches that rely on Metropolis Hastings sampling, or simply estimate the MLE. } 

\strike{Ideally we would want to adapt our approximation to continuous parameter spaces. However, we stress that the goal in this section of the paper was to demonstrate that inferring situational confidence matters, and not necessarily to develop another efficient inverse reinforcement learning algorithm.}

\strike{Lastly, in our user study, we used a single feature change for simplicity and clarity, but our case studies show that the method can be reliably applied to larger feature spaces. However,}

\change{Lastly, our experiments for both demonstrations and corrections are limited to a simple motion planning task with a cost function that depends on only a few features. We do not show how the method would degrade, both under ideal as well as under approximate inference.}

In subsequent work, we hope to address some of these limitations. We are also interested in \strike{extending our work}\change{an extension} to sequential time-dependent inputs, where the person could change their mind about what objective is important to them. Additionally, we want to explore ways of handling misspecification other than reducing learning, such as switching to a more expressive hypothesis space (but demanding more data and computation) whenever the situational confidence is very low for all $\weight$s. Finally, we are excited to showcase our work on other coupled dynamical systems, such as autonomous vehicles.


%

\appendices
\section{\change{Practical Considerations}}
\label{app:practical_considerations}

\subsection{\change{Demonstrations}}
\label{app:practical_demos}

\subsubsection{\change{Discretizing $\wspace$ and $\mathcal{B}$ in \eqref{eq:posterior_demo}}}

\change{For the $\wspace$ discretization, we chose vectors in the unit sphere, as discussed in Section \ref{sec:formulation_cost}. For practical purposes, we restricted the $\weight$ components to be positive due to our task features and the capabilities of our trajectory optimizer; in general, learning from demonstrations should be restricted to norm 1, not necessarily to the positive quadrant. In both our examples in Section \ref{sec:demonstrations} and experiments in Sections \ref{sec:case_study}, each $\weight_i$ component was allowed to take values 0, 0.5, or 1. Since we used 3 features, $\weight$'s dimensionality was 3, leading to a possible set $\wspace$ equivalent to the 3-fold Cartesian product of the values above. After normalizing to norm 1, we were left with 19 unique $\weight$ vectors in $\wspace$, weighing the three features in different proportions, as shown in Figures \ref{fig:simulated_demo}, \ref{fig:case1_easy}, \ref{fig:case1_misspecified}, \ref{fig:case2_correlation}, and \ref{fig:case3_engineering_demo}. Our discretization scheme ensured an approximately uniform sampling on the positive quadrant of the unit sphere.}

\change{To discretize situational confidence, we found it sufficient to cover $\beta\in \{0.01, 0.03, 0.1, 0.3, 1.0, 3.0, 10.0, 30.0, 100.0\}$, the log-scale space, similarly to \cite{fridovich-keil2019confidence,fisac2018probabilistically}. For different tasks, a similar discretization should suffice because what matters is $\beta$'s relative magnitude for identifying misspecification, not its absolute one.
We suggest calibrating the threshold $\epsilon$ in \eqref{eq:threshold_behavior} using a few simulated trajectories like the ones in Fig. \ref{fig:simulated_demo}.}

\subsection{\change{Corrections}}
\label{app:practical_corrections}

\subsubsection{\change{Planning and Replanning}}

\change{We use TrajOpt \cite{trajopt} to plan and replan robot trajectories. We set up the trajectory optimization problem to plan a path that minimizes a cost function of the form of \eqref{eq:phri_cost}. Given different features $\trajfeat$ and weights $\weight$ on these features, different optimal paths may be found. Additionally, we constrain the optimization to plan a path between a pre-specified start and goal locations, while avoiding collisions with the objects in the environment (table, laptop, or human). The total time of the trajectory is fixed, but the actual length can differ. That means that the robot moves faster for longer trajectories, and slower for shorter ones. }

\change{When the experiment starts, the robot plans an initial path from start to goal, using the initial weights $\weight$. When a human push happens, the robot measures the instantaneous deviation, which deforms the trajectory via the impedance controller. Without learning, the robot would resume tracking its original trajectory. However, we use the human input to update $\weight$ according to \eqref{eq:theta_update}, which the robot's planner uses to compute a new trajectory that the robot can follow instead. In a perfect world, this entire process would happen at 60Hz. In practice, however, the trajectory optimizer's computation lasts longer. As such, once a push is registered, the robot starts listening for following torque signals only after the update is complete.}

\change{Imagine this process in the context of a typical user experience. Once the person begins pushing, the robot instantly starts updating $\theta$ and optimizing the new induced path. While the person is applying their correction, the planner eventually finishes its computation and passes the updated trajectory to the robot controller. The user can immediately feel that the robot changed course and stops intervening.}

\subsubsection{\change{Solving \eqref{opt:optimal_uH}}}

\change{We used SLSQP, an off-the-shelf sequential quadratic programming package~\cite{Kraft1988ASP}, to solve \eqref{opt:optimal_uH}. In practice, the method can fail to return a good result if the initialization is bad. We found that if we initialize the minimization with a guess that does not satisfy the constraint (e.g. 0), it returns a reasonable estimate of the true $u_H^*$.}

\subsubsection{\change{Sensitivity Analysis}}

\change{Both \eqref{eq:beta_map} and \eqref{eq:theta_update} rely heavily on hyperparameters $\lambda$ and $\nu$. Here, we discuss how to set them.}

\change{Setting $\lambda$ affects the magnitude of the resulting estimated situational confidence $\hat\beta$ in \eqref{eq:beta_map}. This magnitude plays an important role when later estimating $\weight$ via \eqref{eq:theta_update} because it affects $P(E \mid \hat\beta)$. However, note that to compute this probability we use $P(\hat\beta \mid E)$, which is an entirely data-driven empirical distribution, where the observed $\hat\beta$ is also computed via \eqref{eq:beta_map}. As such, we are not relying on absolute magnitudes of the estimated situational confidence but on relative ones. Therefore, the choice of the hyperparameter $\lambda$ does not affect our method's estimates as long as they are computed with the same hyperparameter that is used for learning $P(\hat\beta \mid E)$.}

\change{In the case of precision $\nu$ in \eqref{eq:bayesnetc_model}, how spread out the Gaussian noise centered around $\trajfeat(\xtraj_R)$ is affects the denominator in \eqref{eq:theta_update}. When $\nu \to 0$, the $\Gamma(\trajfeat_D, E=0)$ term in the denominator goes to 0, which means that \eqref{eq:theta_update} reduces to \eqref{eq:corl_update}: our method always learns and never identifies misspecification. On the other hand, when $\nu \to \infty$, we can use the L'Hospital rule to see that $\Gamma(\trajfeat_D, E=0) \to 0$ as well, as long as $||\trajfeat_D-\trajfeat(\xtraj_R)||^2 \neq 0$, which is true unless there is no correction to deform $\xtraj_R$, in which case we do not need to update $\weight$ at all. Therefore, it is important that $\nu$ is set not too high and not too low in order for our method to work properly. }

\change{The best practice for setting $\nu$ also involves using the offline data calibration from Section \ref{sec:case_study_corrections}. To calibrate properly, after computing the empirical $P(\hat\beta \mid E)$ distribution, when $E = 0$ the updated $\weight$ should not change much, whereas when $E = 1$ the $\weight$ parameter should change appropriately.}

\change{Without the offline data calibration in Section \ref{sec:case_study_corrections}, both $\lambda$ and $\nu$ affect the $\weight$ and $\beta$ estimation, and can have profound effects on the efficacy of our method. Unfortunately, we cannot do this calibration automatically yet, which is a limitation of our work, and we leave it for future research.}

\subsubsection{\change{Trajectory Deformation Parameter Choice}} 
\change{When deforming the robot's trajectory given a human interaction, there are many choices of the deformation matrix $A$ and the deformation magnitude parameter $\mu$. $A$ can be an explicit design choice (for example, constructing $A$ from a finite differencing matrix \cite{bajcsy2017phri}), can be solved for via an optimization problem which penalizes the undeformed trajectory's energy, the work done by the trajectory deformation to the human, and variation’s total jerk as in \change{\cite{losey2017trajectory}}, or can even be learned from human data \cite{jeon2018configuration}. The magnitude of the deformation $\mu$ can also be tuned for best performance, for example to be robust to the rate at which deformations occur (see \cite{losey2018including} for more details).}
\section{Laplace Approximation in Equation \eqref{eq:bayesnetb_model}}
\label{app:laplace}

Let the cost function in the model in \eqref{eq:bayesnetb_model} be denoted by:
\begin{equation}
    \cost_{\change{\trajfeat_D}} (\bar\control) = \lambda\|\bar\control\|^2 + \kappa \|\trajfeat(\bar\xtraj_D) - \change{\trajfeat_D}\|^2,
\end{equation}
for an observed \change{$\trajfeat_D$.}

First, our cost function can be approximated to quadratic order by computing a second order Taylor series approximation about the optimal human action $\control^*_H$ (obtained via the constrained optimization in \ref{opt:optimal_uH}):
\begin{equation}
\begin{split}
    \cost_{\change{\trajfeat_D}}(\bar{\control}) &\approx \cost_{\change{\trajfeat_D}}(\control^*_H) + \nabla \cost_{\change{\trajfeat_D}}(\control^*_H)^{\top}(\bar{\control} - \control^*_H) \\ &+ \frac{1}{2}(\bar{\control} - \control^*_H)^{\top}\nabla^2 \cost_{\change{\trajfeat_D}}(\control^*_H)(\bar{\control} - \control^*_H)\enspace.
\end{split}
\end{equation}
Since $\nabla \cost_{\change{\trajfeat_D}}(\bar{\control})$ has a global minimum at $\control^*_H$ then $\nabla \cost_{\change{\trajfeat_D}}(\control^*_H) = 0$ and the denominator of Equation \ref{eq:bayesnetb_model} can be rewritten as:
\begin{equation}
\begin{split}
\int_\uset &e^{-\beta \cost_{\change{\trajfeat_D}}(\bar{\control})} d\bar{\control} \approx \\ &e^{-\beta \cost_{\change{\trajfeat_D}}(\control^*_H)}\int_\uset e^{-\frac{1}{2}(\bar{\control} - \control^*_H)\beta\nabla^2 \cost_{\change{\trajfeat_D}}(\control^*_H)(\bar{\control} - \control^*_H)} d\bar{\control} \enspace.    
\end{split}
\end{equation}

Since $\beta\nabla^2 \cost_{\change{\trajfeat_D}}(\control^*_H) > 0$ for $\control_H^* \neq 0$, the integral is in Gaussian form, which admits a closed form solution:
$$
\int_\uset e^{-\beta \cost_{\change{\trajfeat_D}}(\bar{\control}_H)} d\bar{\control}_H \approx e^{ -\beta\cost_{\change{\trajfeat_D}}(\control^*_H)} \sqrt{\frac{2\pi^k}{\beta^k|H_{\control^*_H}|}} \enspace,
$$
where $H_{\control^*_H} = \nabla^2 \cost_{\change{\trajfeat_D}}(\control^*_H)$ denotes the Hessian of $\cost_{\change{\trajfeat_D}}$ at $\control^*_H$. Replacing $\cost_{\change{\trajfeat_D}}(\bar{\control}_H)$ with the expanded cost function, we arrive at the final approximation of the observation model:
\begin{align}
	P(\control_H^t &\mid \state^0, \utraj_R ; \Phi_D, \beta) \approx \notag
	\\&  \frac{e^{-\beta\lambda(  \|u_H^{\change{t}}\|^2)}}{ e^{-\beta(\lambda \|{\control_H^*}\|^2 + \change{\kappa}\|\trajfeat(\xtraj_D^*) - \trajfeat_D\|^2)}}\sqrt{\frac{\beta^k|H_{\control^*_H}|}{2\pi^k}}
    \enspace.
\end{align}


\section*{Acknowledgment}

This research is supported by the Air Force Office of Scientific Research (AFOSR) and the Open Philanthropy Project.

\ifCLASSOPTIONcaptionsoff
  \newpage
\fi


{
\bibliographystyle{IEEEtran}
\bibliography{IEEEabrv,TRO2020}}
%



%

\begin{IEEEbiography}[{\includegraphics[width=1in,height=1.25in,clip,keepaspectratio]{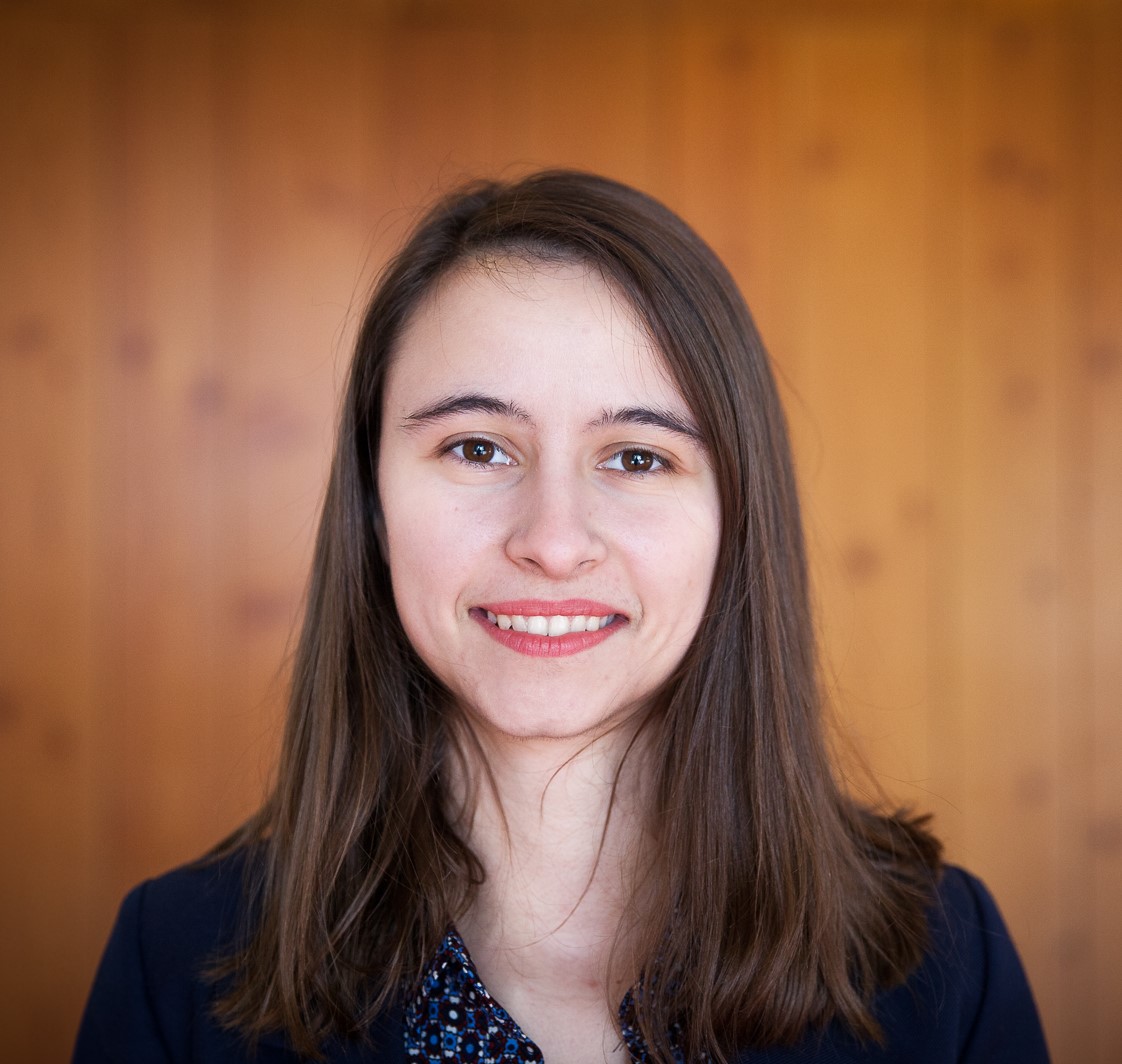}}]{Andreea Bobu} received her B.S. degree in computer science and engineering from the Massachusetts Institute of Technology, United States in 2017. She is currently a Ph.D. student with the Interactive Autonomy and Collaborative Technologies Laboratory at University of California Berkeley in the United States. Her research interests include machine learning techniques for robot learning with uncertainty.
\end{IEEEbiography}

\begin{IEEEbiography}[{\includegraphics[width=1in,height=1.25in,clip,keepaspectratio]{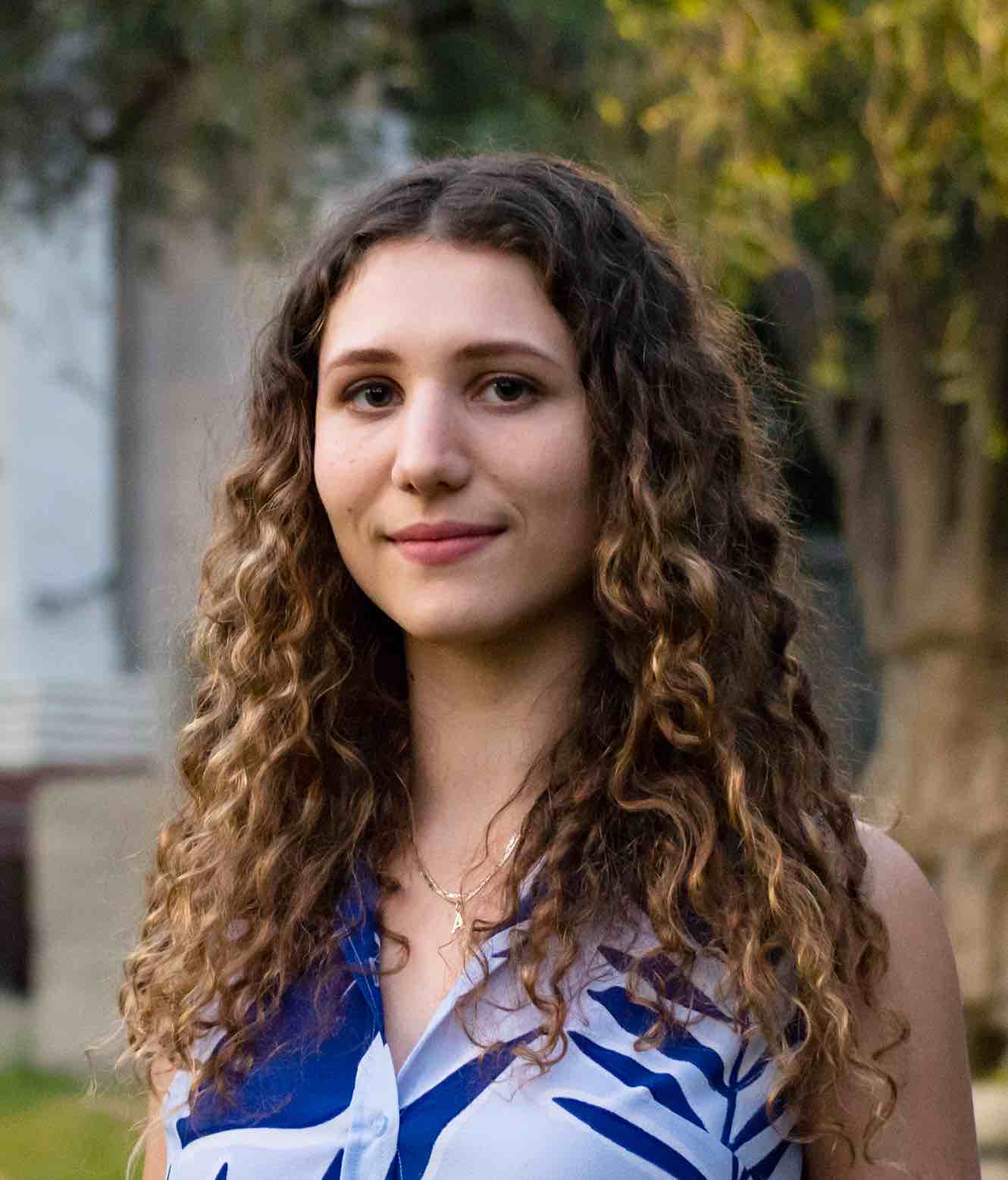}}]{Andrea Bajcsy} received her B.S. degree in computer science from the University of Maryland College Park, United States in 2016. She is currently a Ph.D. student in Electrical Engineering and Computer Sciences at the University of California Berkeley in the United States. She is the recipient of the National Science Foundation's Graduate Research Fellowship. 
\end{IEEEbiography}

\begin{IEEEbiography}[{\includegraphics[width=1in,height=1.25in,clip,keepaspectratio]{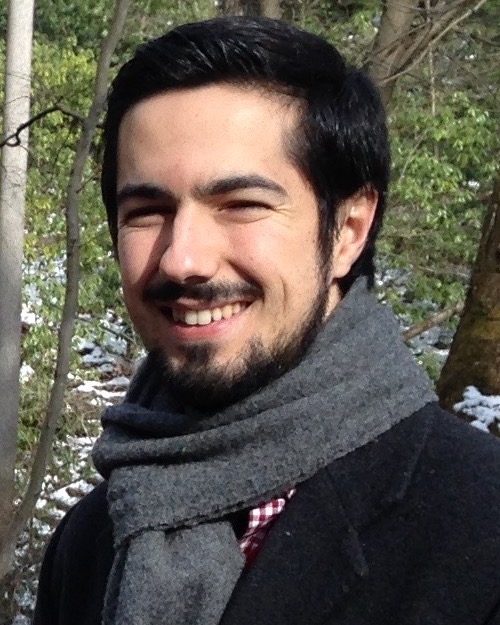}}]{Jaime F. Fisac}
received a B.S./M.S. degree in Electrical Engineering from the Universidad Politécnica de Madrid, Spain, in 2012, the M.Sc. degree in Aeronautics from Cranfield University, U.K., in 2013, and the Ph.D. degree in Electrical Engineering and Computer Sciences from the University of California, Berkeley, U.S.A. in 2019. His research interests lie in control theory and artificial intelligence, with a focus on safety for autonomous systems. He is a recipient of the ``la Caixa" Foundation Fellowship.
\end{IEEEbiography}


\begin{IEEEbiography}[{\includegraphics[width=1in,height=1.25in,clip,keepaspectratio]{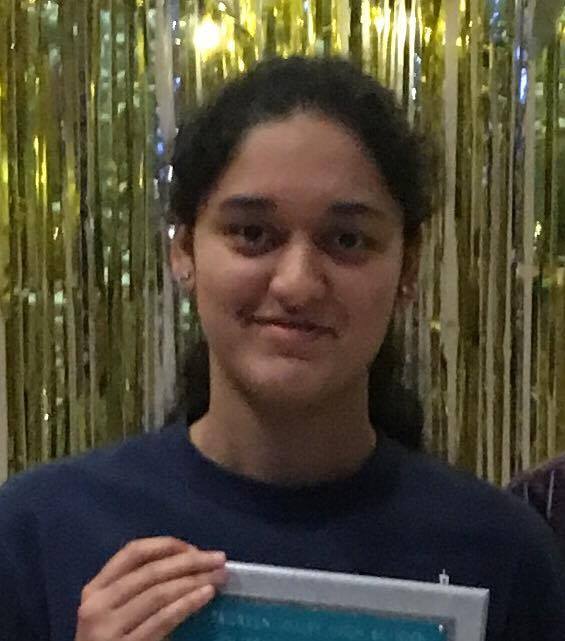}}]{Sampada Deglurkar} is a third year undergraduate student working towards her B.S. degree in Electrical Engineering and Computer Sciences at the University of California, Berkeley in the United States.
\end{IEEEbiography}

\begin{IEEEbiography}[{\includegraphics[width=1in,height=1.25in,clip,keepaspectratio]{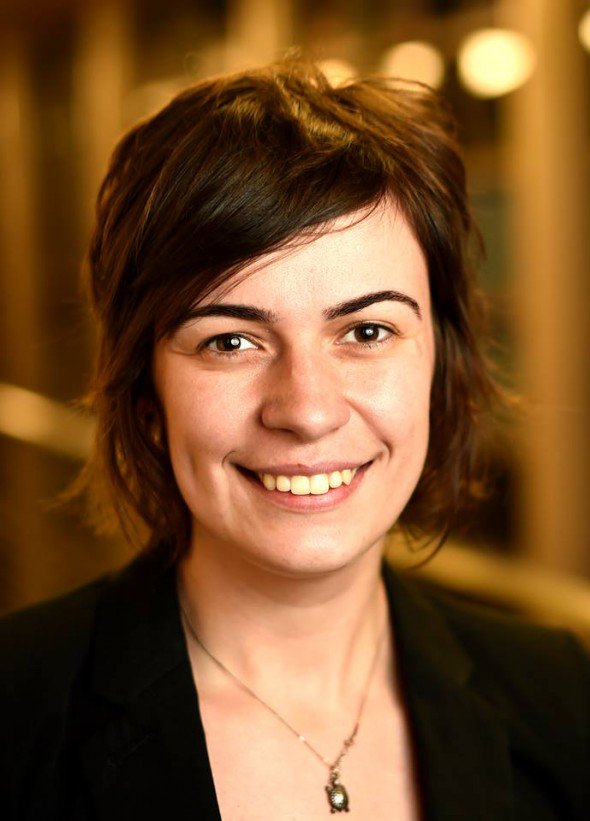}}]{Anca Dragan} received her Ph.D. degree in robotics from Carnegie Mellon University in 2015. She is currently an Assistant Professor in the Electrical Engineering and Computer Science department at the University of California Berkeley in Berkeley, United States.

\end{IEEEbiography}


\vfill


\end{document}